\documentclass[11pt]{article}

\usepackage[preprint]{acl}

\usepackage{times}
\usepackage{latexsym}

\usepackage[T1]{fontenc}

\usepackage[utf8]{inputenc}

\usepackage{microtype}

\usepackage{inconsolata}

\usepackage{graphicx}

\usepackage{subfig}
\usepackage{graphicx}
\usepackage{amsmath}
\usepackage{amssymb}
\usepackage{booktabs}
\usepackage{multirow}
\usepackage{xcolor}
\usepackage{tabularx} 
\usepackage{booktabs} 
\definecolor{square_bracket_color}{HTML}{0E8088} 
\definecolor{curly_bracket_color}{HTML}{B85450} 
\definecolor{red_apple}{HTML}{FF0000} 
\definecolor{purple_apple}{HTML}{9933FF} 
\usepackage{pifont} 
\usepackage{diagbox}
\newcommand{\twoshotsecond}[1]{{\textcolor{blue}{\underline{#1}}}}
\newcommand{\twoshotfirst}[1]{{\textbf{\textcolor{red}{#1}}}}
\newcommand{\fourshotfirst}[1]{{\textbf{\textcolor[HTML]{FF8000}{#1}}}}
\newcommand{\fourshotsecond}[1]{{\textcolor[HTML]{008080}{\underline{#1}}}}

\title{Think Bright, Diffuse Nice: Enhancing T2I-ICL via Inductive-Bias Hint Instruction and Query Contrastive Decoding}

\author{
 \textbf{Zhiyong Ma\textsuperscript{1,2}},
 \textbf{Zhenpeng Li\textsuperscript{1}},
 \textbf{Yuanjie Shi\textsuperscript{3}}
 \textbf{Zhengping Li\textsuperscript{4}},
 \textbf{Jiahao Chen\textsuperscript{1}},
 \textbf{Qingyuan Chuai\textsuperscript{1}},
\\
\\
 \textsuperscript{1}Cao Tu Li (Guangzhou) Technology Co., Ltd, Guangzhou, Guangdong, China,\\
 \textsuperscript{2}South China University of Technology, Guangzhou, Guangdong, China,\\
 \textsuperscript{3}Washington State University, Pullman, Washington State, United States of America,\\
 \textsuperscript{4}Hong Kong Baptist University, Kowloon, Hong Kong
\\
 \small{
   \textbf{Correspondence:} \href{qingyuanchuai@outlook.com}{qingyuanchuai@outlook.com}
 }
}

\begin{document}
\maketitle
\begin{abstract}
Text-to-Image In-Context Learning (T2I-ICL) enables customized image synthesis via interleaved text-image examples but faces two mutually reinforcing bottlenecks, compliance failure and prior-dominated hallucination, that form a vicious cycle degrading generation quality.
Existing methods rely on tailored training, which limits flexibility and raises deployment costs. 
To address these challenges effectively, we propose TBDN, a training-free framework integrating two complementary closed-loop mechanisms: Hint Instruction (HI) and Query Contrastive Decoding (QCD).
HI injects task-aware inductive bias via lightweight prompt engineering to anchor models on contextual mapping rules, thereby mitigating compliance failure.
QCD adjusts the decoding distributions of language models by contrasting full-input and query-omitted distributions, suppressing prior-dominated hallucination. 
TBDN achieves State-of-the-Art performance on CoBSAT and Text-to-Image Fast Mini-ImageNet, with robust generalization across model backbones, prompt designs, and hyperparameters.
It also maintains promising performance in concept preservation and prompt following on Dreambench++. 
By breaking the two bottlenecks, TBDN establishes a simple yet effective framework for efficient and reliable T2I-ICL.
\end{abstract}

\section{Introduction}
\label{sec:intro}
Diffusion models have emerged as the mainstream paradigm in image generation, with numerous related methods proposed and widely deployed in educational and industrial settings~\cite{diffusion_survey}.
These methods generate corresponding visual outputs from either a single textual prompt or a text-image pair.
Yet certain concepts requiring visualization are difficult to articulate adequately via these inputs.
For example, a designer may need to generate a “desert-themed cow” (combining two unrelated concepts) for a marketing campaign, or a teacher may want to visualize a “purple watercolor apple” (modifying attributes and style) to explain color theory.
To convey such concepts, humans often rely on a series of interleaved text-image examples (referred to as contexts~\cite{mmicl_survey} or demonstrations~\cite{thinkdiff}).
Addressing this visualization need requires methods to adapt to such complex input patterns and perform semantic reasoning over the input content to produce coherent results, a research challenge formally defined as the Text-to-Image In-Context Learning (T2I-ICL) task~\cite{cobsat}.
\begin{figure*}[htbp]
    \centering
    \includegraphics[width=0.98\textwidth]{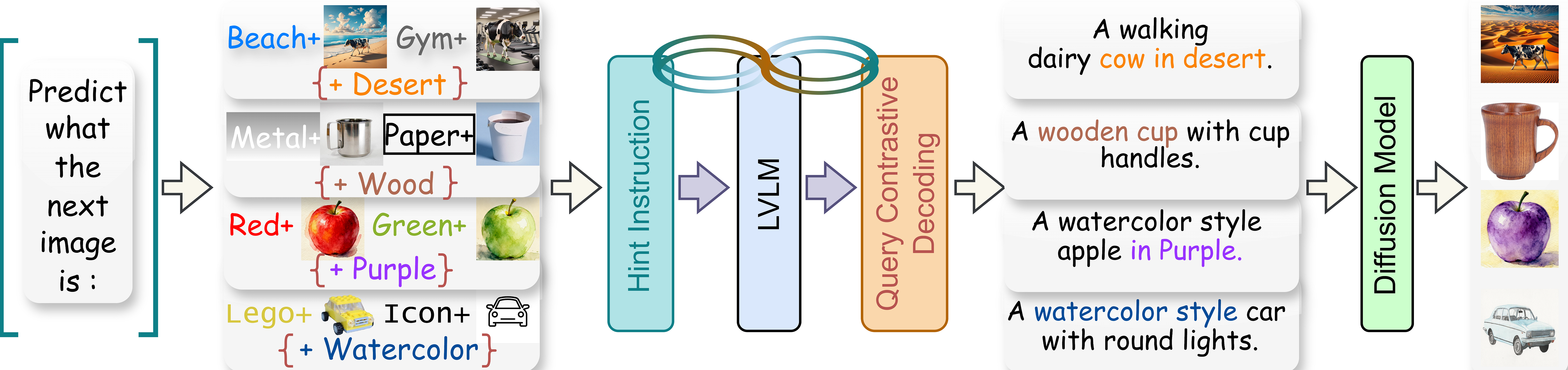}
    \caption{
    Overview of the TBDN framework for T2I-ICL. 
    Given an instruction (enclosed in \textcolor{square_bracket_color}{`[~]'}), context examples, and queries (enclosed in \textcolor{curly_bracket_color}{`(~)'}), TBDN first injects task-aware inductive bias via Hint Instruction, then refines outputs via Query Contrastive Decoding to suppress priors, and finally generates images with a diffusion model.
    }
    \label{fig: Overview}
\end{figure*}

Existing unified multimodal large language models (unified MLLMs) exhibit in-context learning (ICL) capabilities comparable to those of large language models (LLMs)~\cite{emu,seed_llama}, while extending such abilities to multimodal understanding and generation within a unified architectural framework~\cite{emu2}.
However, recent studies~\cite{cobsat, vlicl} have revealed that unified MLLMs struggle to effectively leverage their reasoning capabilities in T2I-ICL without tuning.
To achieve robust reasoning and higher-fidelity generation results, another line of work~\cite{thinkdiff} adopts an intuitively effective paradigm: integrating large vision-language models (LVLMs) with diffusion models.
This paradigm aims to harness the strong reasoning capabilities of LVLMs alongside the superior fidelity and semantic controllability of diffusion models to attain desired T2I-ICL performance.

Specifically, LVLMs first process the interleaved contexts to extract semantic rules, and their output $y$ is then fed into diffusion models for image generation.
Methods under this paradigm fall into two categories: (1) injecting the latent representations of $y$ into visual decoder for decoding~\cite{emu, thinkdiff}; (2) feeding the textual outputs (or tokens in $y$) into the image generator to complete encoding and decoding~\cite{imagegen_cot}. 
The former achieves unified input processing but often suffers from representation drift, requiring further training and incurs substantial alignment costs.
By contrast, the latter is more direct and interpretable, yet relevant explorations remain scarce and lack systematic design principles.

Recalling that the core requirement of T2I-ICL is that, given a few demonstrations, the method should infer underlying mapping rules, generate corresponding descriptive guidance, and perform visualization accordingly. 
However, both literature and our empirical analysis reveal two key bottlenecks that degrade performance: \textbf{($\dagger$) Compliance failure}; \textbf{($\dagger\dagger$) Prior-dominated hallucination}.
These bottlenecks form a vicious cycle.
When a method fails to discern mapping rules, it relies more on prior knowledge to satisfy generation demands. 
In turn, prior-dominated generation consumes attentional resources~\cite{marinescu2025relationshipchoicerepresentationincontext}, further hindering the method from mining contextual mapping rules and exacerbating compliance failures.

To break this cycle and address two bottlenecks effectively, we propose \textbf{TBDN} (Fig.~\ref{fig: Overview}), a textual-output-driven framework integrating two complementary, mutually reinforcing mechanisms that form a \textit{Möbius Band}-like closed-loop constraint:
\begin{itemize}
    \item \textbf{Hint Instruction~(HI)}: A prompt engineering strategy that injects task-aware inductive bias to resolve compliance failure (Fig.~\ref{fig: HI}).
    \item \textbf{Query Contrastive Decoding~(QCD)}: A decoding approach that imposes posterior instruction-following constraints to eliminate prior-dominated hallucination (Fig.~\ref{fig: QCD}).
\end{itemize}
HI appends a priori guidance to the input instruction, directing the LVLM to prioritize the input query.
It anchors the model’s focus on mapping rules rather than superficial mimicry, addressing compliance failure at the root via targeted inductive bias.
Complementing HI, QCD refines generation distributions to suppress prior over-reliance and amplify query-aligned knowledge, neutralizing hallucination by aligning output with contextual rules.

To validate the effectiveness of TBDN, we conduct comprehensive experiments across benchmarks and shot settings, where it achieves state-of-the-art performance in most cases.
It further demonstrates strong generalization across LVLM backbones, with pronounced performance gains from the synergistic integration of HI and QCD.
Most notably, TBDN is training-free, outperforming existing methods relying on cumbersome training while maintaining superior performance.

Our contributions in T2I-ICL are as follows:
\begin{itemize}
    \item We identify two mutually exacerbating bottlenecks (compliance failure, prior-dominated hallucination) and their vicious cycle, clarifying a principled method design direction.
    \item We propose TBDN\footnote{\url{https://github.com/Calendula597/TBDN}}, a training-free framework with two complementary, closed-loop mechanisms (HI / QCD) that resolve these two bottlenecks respectively.
    \item TBDN achieves strong cross-benchmark performance: it reaches State-of-the-Art on CoBSAT and Text-to-Image Fast Mini-ImageNet, with robust generalization across LVLM backbones, prompt designs and hyperparameters. 
    On Dreambench++, it further demonstrates promising prompt-following capability.
\end{itemize}

    


\section{Related Work}
\label{sec:related}
\subsection{Diffused Image Generation}
\label{sec:related:dig}
Multimodal image generation aims to generate images conditioned on textual descriptions and, optionally, reference images. 
As promising methods in this field, diffusion models excel at reconstruction fidelity.
Representative methods such as DALL-E~\cite{dalle} and Stable Diffusion~\cite{sd,sdxl} leverage language models to process textual inputs, laying the semantic foundation for downstream generation. 
To control the structural and stylistic attributes, approaches such as ControlNet~\cite{controlnet}, T2I-Adapter~\cite{t2i_adapter}, and FLUX~\cite{flux} incorporate multimodal encoders to capture semantic information from inputs. 
Beyond text-only or image-text pair inputs, interleaved inputs, which contain multiple images and text, offer richer and more faithful representations of user intention. 
Recent approaches have attempted to interpret and leverage them for image generation~\cite{rpg, tf_ti2i}.

\subsection{Text-to-Image In-Context Learning} 
\label{sec:related:icl}
LLMs~\cite{icl,wei2023largerlanguagemodelsincontext}, LVLMs~\cite{llava,flamingo,qwen2vl}, and MLLMs~\cite{gill,emu} have exhibited remarkable ICL performance~\cite{icl_survey}.
To unify image generation with ICL, CoBSAT~\cite{cobsat} formalizes the T2I-ICL task.
This work not only provides a valuable benchmark but also demonstrates that fine-tuning and prompt engineering are effective strategies for language models in this task. 
Building on the fine-tuning paradigm, ThinkDiff~\cite{thinkdiff} trains an aligner network to transfer multimodal in-context reasoning capabilities from VLMs to diffusion models. 
It conducts a captioning training task to align the representation space of two foundational modules (VLM and diffusion decoder). 
Integrating these insights, researchers~\cite{imagegen_cot} introduce a chain-of-thought dataset (ImageGen-CoT) and fine-tune SEED-X~\cite{seedx} and SEED-LLaMA~\cite{seed_llama} on it, encouraging language models to generate textual analysis prior to image generation.
Despite their effectiveness, these methods typically rely on significant data and computational resources, which restricts their deployment.

\section{Two Bottlenecks in T2I-ICL}
\label{sec:Bottlenecks}
\begin{figure}[htbp]
\centering
\includegraphics[width=0.49\textwidth]{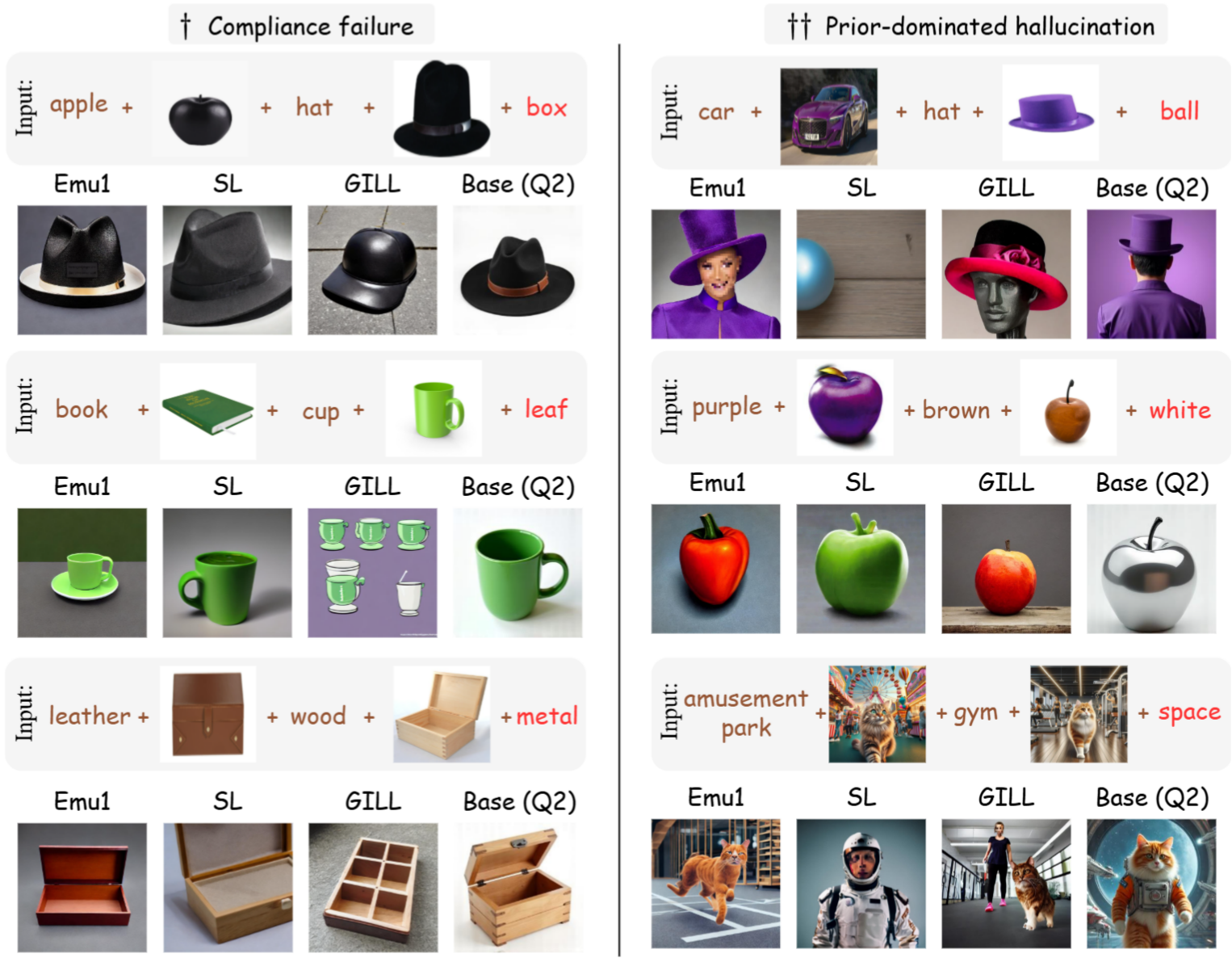}
  \caption{Two critical bottlenecks in T2I-ICL (evaluated on CoBSAT).
  Compliance failure (left): methods parrot input context (e.g., ``hat'', ``cup'') instead of reasoning query semantics. 
  Prior-dominated hallucination (right)  methods generate prior-aligned outputs (e.g., ``red/green apples'') that violate input requirements.}
  \label{fig:two_problem}
\end{figure}
Our empirical practice shows many T2I-ICL methods fail to recognize mapping rules in contexts and instead capture superficial features, which manifests as context parroting, a mechanical repetition of input context~\cite{zhang2025contextparrotingsimpletoughtobeat}.
We term this phenomenon \textbf{Compliance failure}.
For methods equipped with language models, strong linguistic and visual priors from pre-trained data carry excessive weight during generation~\cite{WhyLargerLanguageModelsDoIncontextLearningDifferently}, suppressing input requirements and inducing prior-aligned yet context-violating outputs.
We term this \textbf{Prior-dominated hallucination}.

Fig.~\ref{fig:two_problem} illustrates representative cases of these two bottlenecks.
As shown on the left, methods repeat concepts in the input context (e.g., parroting ``hat'', ``cup'', and ``wood'') rather than reasoning according to the input query.
On the right, methods exhibit obvious prior-dominated hallucination: concepts like ``hat'' and ``space'' are associated with ``human'', while ``apple'' (or other objects of similar shape) is associated with the ``red'' and ``green'' attributes.
Such issues can be further exemplified, and we present more illustrative cases in the appendix.

\begin{figure}[htbp]
\centering
\includegraphics[width=0.48\textwidth]{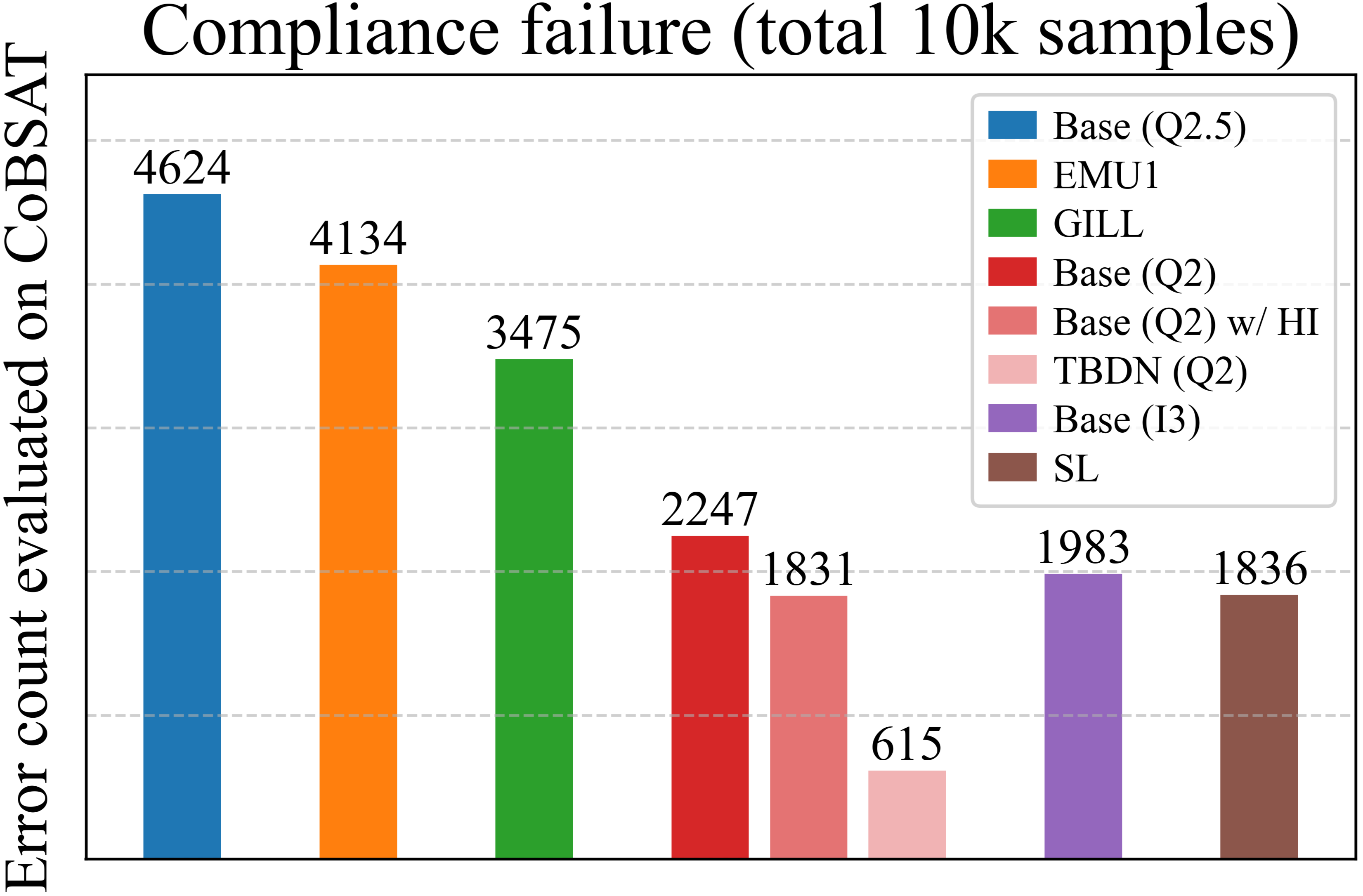}
  \caption{Compliance failure error counts across methods on CoBSAT.}
  \label{fig:compliance_failure_bar}
\end{figure}

Prior-dominated hallucination is intuitive and prevalent, while compliance failure is relatively obscure.
To clarify this concept and formally measure it, we draw on the framework of CoBSAT, which evaluates model responses along two dimensions (object and attribute) and define an error count metric.
This metric counts the number of samples that satisfy at least one of the following conditions: (1) the predicted attribute matches the ground-truth attribute, while the predicted object appears in the input context; (2) the predicted object matches the ground-truth object, while the predicted details appear in the input context.
Using this metric, we quantify the error counts of different methods over 10k samples on CoBSAT, revealing consistent failure patterns (Fig.~\ref{fig:compliance_failure_bar}). 

\section{Method}
\label{sec:method}
\subsection{Overview}

\begin{figure}[htbp]
    \centering
    \includegraphics[width = 0.48\textwidth]{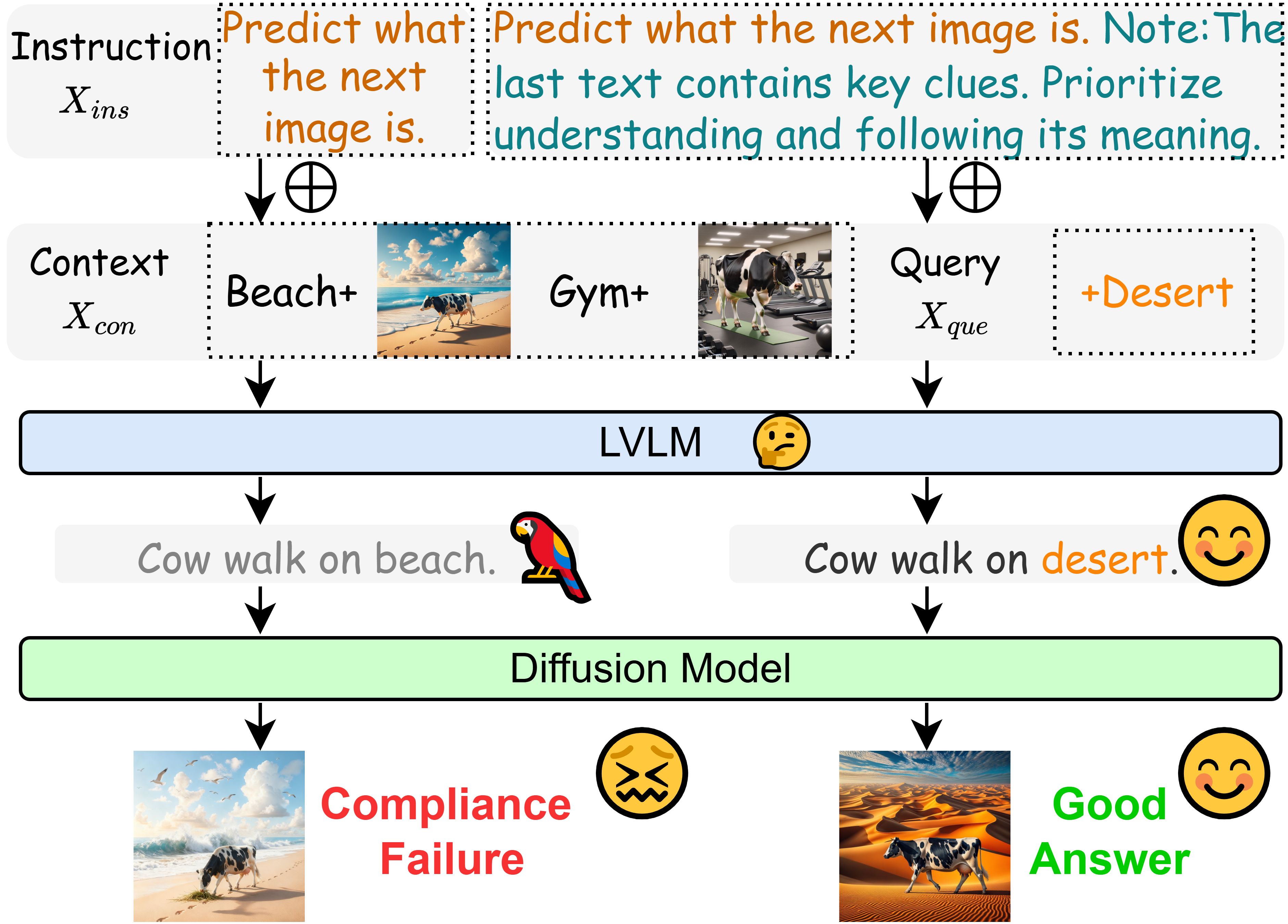}
    \caption{
    Overview of Hint Instruction, a mechanism guiding LVLMs to prioritize context-aware query reasoning, thereby mitigating compliance failure.
    }
    \label{fig: HI}
\end{figure}

\begin{figure}[htbp]
    \centering
    \includegraphics[width = 0.49\textwidth]{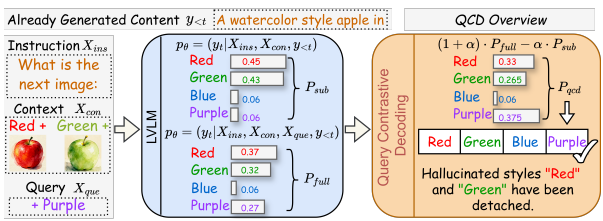}
    \caption{
    Overview of QCD which adjusts decoding distributions to mitigate prior-dominated hallucinations.
    }    
    \label{fig: QCD}
\end{figure}

TBDN is a simple yet effective training-free T2I-ICL framework unifying LVLMs and diffusion models, addressing two core bottlenecks: compliance failure and prior-dominated hallucination.
Following the ``Think Bright, Diffuse Nice" philosophy, it employs two non-redundant mechanisms, Hint Instruction and Query Contrastive Decoding, to form a closed loop for rule alignment and prior suppression.
The workflow of TBDN has five stages: 
(1) Pre-processing: instruction $X_{\text{ins}}$, interleaved text-image context $X_{\text{con}}$, and query $X_{\text{que}}$ are concatenated as a unified multimodal sequence; 
(2) Injection: hint prompt is added at the end of $X_{\text{ins}}$ to anchor contextual mapping rules (Fig.~\ref{fig: HI}); 
(3) Reasoning: the injected input is fed into the LVLM and produce corresponding distribution; 
(4) Decoding: two distinct distributions $P_{\text{sub}}$ (conditioned on instruction and contexts) and $P_{\text{full}}$ (conditioned on full input) are used to perform weighted contrastive adjustment (Fig.~\ref{fig: QCD}); 
(5) Diffusion: the LVLM’s rule-aligned textual output is passed to the diffusion model for high-fidelity generation.

\subsection{Hint Instruction}
\label{sec:method:hint}
Prior works~\cite{cobsat,vlicl} have observed that LVLMs struggle to reliably comprehend input semantics and might respond irrelevantly in T2I-ICL. 
In our empirical observations, we further find that LVLMs tend to disregard input queries and instead recapitulate or even exhibit context parroting~\cite{zhang2025contextparrotingsimpletoughtobeat}.
We term such issues collectively as compliance failure.
To address compliance failure, tailored training might be effective but incurs prohibitive computational and data costs~\cite{imagegen_cot}.
By contrast, prompt engineering~\cite{prompt_survey,least_to_most_prompting} offers another route by injecting external inductive bias. 
Previous works~\cite{cobsat,thinkdiff} have explored this direction but have failed to achieve satisfactory performance.
For instance, Chain-of-Thought (CoT)~\cite{cot} delivers marginal improvements but increases inference cost and risks exceeding length limits.

For effectiveness and efficiency, we propose Hint Instruction (HI), a prompt-based strategy illustrated in Fig.~\ref{fig: HI}.
Unlike related works~\cite{cobsat,thinkdiff,imagegen_cot} that focus on context expansion, HI injects a task-aware inductive bias to directly align the model’s reasoning with query semantics. 
Specifically, based on prior findings~\cite{li2025makelvlmsfocuscontextaware} and empirical evaluation, LVLMs tend to ignore or misunderstand the input query, which is typically positioned at the end of the input sequence. 
This oversight leads to suboptimal performance in T2I-ICL, as the model fails to prioritize the query’s guiding role. 
To counter this issue, we reverse the flaw into a targeted inductive bias: \textbf{the query is a vital cue and should take precedence}.
Guided by this core bias, we define two design principles for HI's prompt construction: ($\ddagger$)~The query provides key guidance for subsequent generation; 
($\ddagger\ddagger$)~The query semantics take precedence even when there is semantic conflict between the query and context.
We implement HI by suffixing an instruction adopted from~\cite{thinkdiff} with the following sequence: 
\begin{quote}
\textit{``The last text I provide contains the most important clue about the next picture. Focus mainly on understanding and following the meaning of the final text when creating your description.''}
\end{quote}
This specific prompt is selected via extensive ablation experiments, which verify its superiority over alternative designs.
Overall, HI augments the LVLM with minimal input overhead by instantiating the aforementioned inductive bias, yielding substantial gains without additional training cost.




\begin{figure}[htbp]
    \centering
    \includegraphics[width=0.48\textwidth]{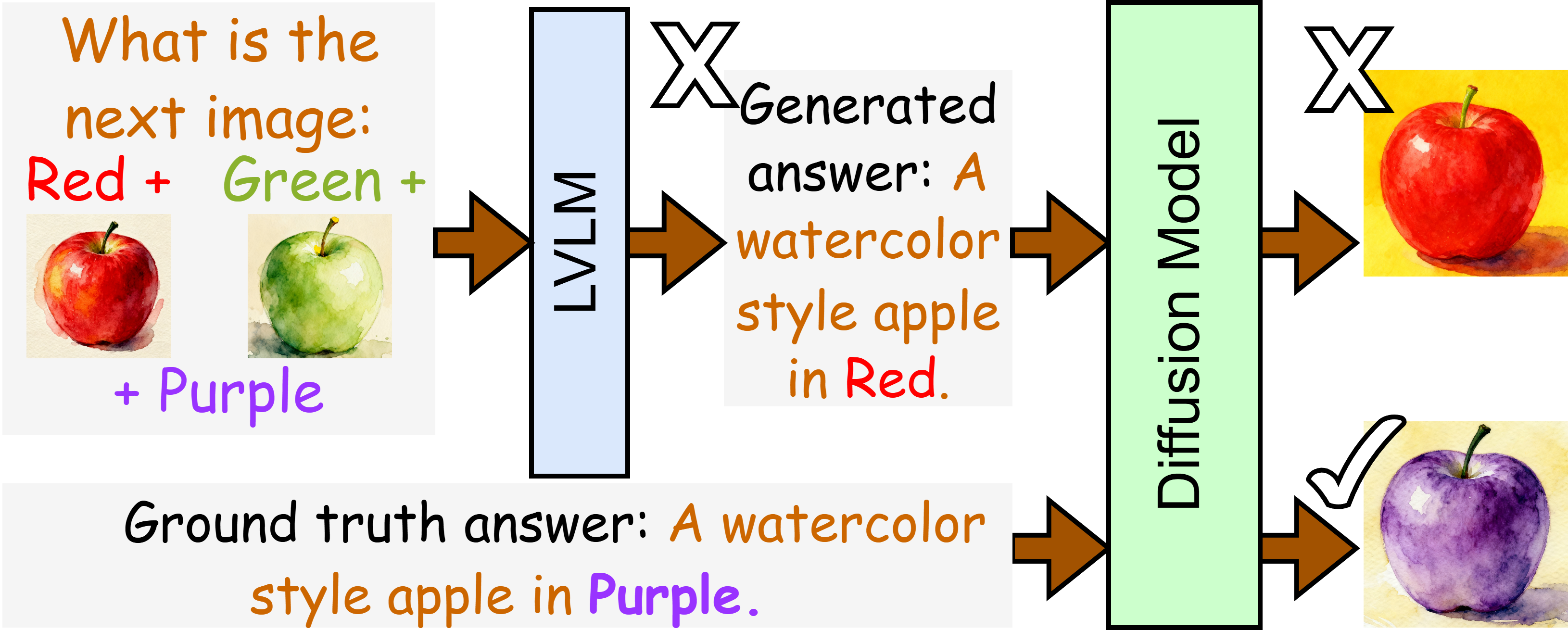}
    \caption{
    An illustrative example of prior-dominated hallucination in LVLMs for T2I-ICL. 
    Given text-image pairs and the final query, the LVLM exhibits reliance on its prior association (apple~$\leftrightarrow$~red) to generate an incorrect description, leading the diffusion model to render a red apple that mismatches the semantics of the ground truth answer. 
    This failure motivates our Query Contrastive Decoding (QCD) method, which mitigates such hallucination by adjusting LVM output distributions.
    }
    \label{fig: Red_Apple_Green_Apple_Purple}
\end{figure}

\subsection{Query Contrastive Decoding}
\label{sec:method:qcd}
Apart from the compliance failure, LVLMs are also usually affected by the prior-dominated hallucination~\cite{niu2021counterfactual,li2023evaluating}.
Fig.~\ref{fig: Red_Apple_Green_Apple_Purple} presents a concrete case of prior-dominated hallucination, where the LVLM incorrectly associates ``apple'' with \textcolor{red_apple}{``red''} instead of analyzing the target query \textcolor{purple_apple}{``purple''}.
To mitigate such problem, we propose Query Contrastive Decoding (QCD) inspired by \cite{VCD,ICD} that adjust the output distributions of LVLMs (Fig. \ref{fig: QCD}).

Given an input instruction $X_{ins}$, context $X_{con}$, and query $X_{que}$, the input of T2I-ICL can be formally denoted as $X = [X_{\text{ins}};X_{\text{con}};X_{\text{que}}]$.
The LVLM (parameterized by $\theta$) is expected to autoregressively generate a description $Y$ of the target image, which is a token sequence length with $L$: $Y = [y_{1};y_{2};\cdots;y_{L}]$.
Conventionally, each token $y_t$ (where 1$\leq$ t$\leq$ L) in $Y$ is sampled from a probability distribution $P_{full}$ during decoding, which is mathematically formulated as:
\begin{gather}
P_{full} = \prod_t p_{\theta}(y_t \mid X_{ins}, X_{con}, X_{que}, y_{<t}),
\end{gather}
where $y_{<t}$ denotes tokens generated up to time step $(t - 1)$.
To implement QCD, we compute a secondary distribution $P_{\text{sub}}$ from the same input with the query omitted, which is given by:
\begin{equation}
P_{sub} = \prod_t p_{\theta}(y_t\mid X_{ins}, X_{con}, y_{<t}).
\end{equation}
Leveraging $P_{sub}$ and $P_{full}$, QCD generates the response $Y$ by sampling from the query contrastive distribution $P_{qcd}$, which amplifies differences between the two distributions via hyperparameter $\alpha$:
\begin{equation}
P_{qcd} = softmax((1 + \alpha) \cdot P_{full} - \alpha \cdot P_{sub}).
\end{equation}
A larger $\alpha$ strengthens this amplification, and $\alpha = 0$ reverts to regular decoding strategy.


\section{Experiments}
\label{sec:experiments}
\begin{table*}[hbtp]
\centering
\resizebox{\textwidth}{!}{
\begin{tabular}{c|ccccc|ccccc|l}
\hline
\multirow{2}{*}{\textbf{Method}}                                                         & \multicolumn{5}{c|}{\textbf{Object-Inference Task}}        & \multicolumn{5}{c|}{\textbf{Attribute-Inference Task}}          & \multirow{2}{*}{\textbf{Avg. acc.}$\uparrow$} \\ \cline{2-11}
                                                                                & Color-I & Bkg.-I & Style-I & Action-I & Texture-I & Color-II & Bkg.-II & Style-II & Action-II & Texture-II &                       \\ \hline
GILL                                                                            & .171    & .054   & .069    & .063     & .074      & .010     & .043    & .024     & .022      & .040       & .057                  \\
Emu1                                                                             & .065    & .051   & .057    & .052     & .078      & .062     & .109    & .081     & .092      & .074       & .072                  \\
SL                                                                      & .616    & .216   & .272    & .592     & .112      & .088     & .168    & .192     & .220      & .056       & .254                  \\
SL-IGC & .620    & .368   & .384    & .424     & .060      & .192     & .288    & .208     & .216      & .148       & .291                  \\
SX                                                                          & .796    & .412   & .316    & .596     & .240      & .176     & .344    & .260     & .252      & .104       & .349                  \\
ThinkDiff                                                                       & .622    & .349   & .237    & .459     & .290      & .511     & .534    & .340     & .534      & .292       & .417                  \\
SX-IGC    & .884    & .692   & \twoshotfirst{.928}    & \twoshotfirst{.936}     & .420      & .504     & .612    & \twoshotfirst{.660}     & .524      & .424       & .658                  \\ \hline
Base~(Q2)
& .766    & .719   & .299    & .498     & .426      & .501     & .769    & .338     & .686      & .366       & .536                  \\
TBDN (Q2)                                                            & \twoshotsecond{.909}    & \twoshotfirst{.879}   & \twoshotsecond{.500}    & .760     & \twoshotfirst{.556}      & \twoshotfirst{.724}     & \twoshotsecond{.905}    & \twoshotsecond{.487}     & .683      & \twoshotsecond{.531}       & \twoshotfirst{.693}($\uparrow$29.2\%)                  \\ \hline
Base~(Q2.5)                                                           & .354    & .407   & .127    & .392     & .223      & .231     & .442    & .213     & .522      & .217       & .312                  \\
TBDN (Q2.5)                                                            & .843    & .594   & .347    & .624     & .459      & .534     & .794    & .409     & .589      & .441       & .563($\uparrow$80.1\%)                  \\ \hline
Base~(I3)                                                           & .826    & .760   & .317    & .711     & .432      & .385     & .866   & .442     & \twoshotsecond{.701}      & .429       & .586                  \\
TBDN (I3)                                                            & \twoshotfirst{.927}    & \twoshotsecond{.808}   & .383    & \twoshotsecond{.808}    & \twoshotsecond{.520}      & \twoshotsecond{.587}     & \twoshotfirst{.956}    & .486     & \twoshotfirst{.789}      & \twoshotfirst{.568}      & \twoshotsecond{.683}($\uparrow$16.4\%)                  \\ \hline
\end{tabular}
}
\caption{Comparison of 2-shot accuracy on CoBSAT. “Bkg.” denotes background, and “Avg. Acc.” is the average accuracy over 10 tasks. \twoshotfirst{Red} and \twoshotsecond{blue} highlight the best and second-best results per task.}
\label{tab:cobsat:2shot}
\end{table*}

\begin{table*}[htbp]
\resizebox{\textwidth}{!}{
\begin{tabular}{c|ccccc|ccccc|l}
\hline
\multirow{2}{*}{\textbf{Method}} & \multicolumn{5}{c|}{\textbf{Object-Inference Task}}        & \multicolumn{5}{c|}{\textbf{Attribute-Inference Task}}          & \multirow{2}{*}{\textbf{Avg. acc.}$\uparrow$} \\ \cline{2-11}
                        & Color-I & Bkg.-I & Style-I & Action-I & Texture-I & Color-II & Bkg.-II & Style-II & Action-II & Texture-II &                                 \\ \hline
GILL                    & .106    & .044   & .041    & .073     & .087      & .022     & .059    & .044     & .032      & .067       & .058                            \\
Emu                     & .063    & .018   & .045    & .048     & .097      & .037     & .122    & .109     & .077      & .088       & .070                            \\
SL              & .482    & .211   & .141    & .053     & .122      & .252     & .076    & .268     & .207      & .105       & .192                            \\
ThinkDiff               & .638    & .362   & .254    & .434     & .317      & .610     & .590    & .432     & .664      & .332       & .463                            \\
\hline
Base~(Q2)                & .815    & .730   & .346    & .466     & .504      & .652     & .920    & .443     & .812      & .449       & .614                            \\
TBDN (Q2)                    & \fourshotsecond{.920}   & \fourshotsecond{.948}   & \fourshotfirst{.582}    & \fourshotsecond{.822}     & \fourshotfirst{.633}      & \fourshotfirst{.851}     & \fourshotsecond{.980}    & \fourshotfirst{.550}     & \fourshotsecond{.788}      & \fourshotsecond{.593}       & \fourshotsecond{.767}($\uparrow$24.9\%)                            \\ \hline
Base~(Q2.5)             & .498    & .558   & .149    & .446     & .255      & .369     & .493    & .265     & .620      & .296       & .395                            \\
TBDN (Q2.5)                    & .868    & .832   & .392    & .752     & .531      & .675     & .960    & .457     & .740      & .517       & .672($\uparrow$70.1\%)                            \\ \hline
Base~(I3)               & .881    & .908   & .401    & .824     & .573      & .659     & .936    & .522     & .837      & .588       & .713                            \\
TBDN (I3)                    & \fourshotfirst{.927}    & \fourshotfirst{.951}   & \fourshotsecond{.490}    & \fourshotfirst{.875}     & \fourshotsecond{.627}      & \fourshotsecond{.783}     & \fourshotfirst{.983}    & \fourshotsecond{.547}     & \fourshotfirst{.868}      & \fourshotfirst{.636}       & \fourshotfirst{.769}($\uparrow$7.8\%)                           \\ \hline
\end{tabular}
}
\caption{Comparison of 4-shot accuracy on CoBSAT. “Bkg.” denotes background, and “Avg. Acc.” is the average accuracy over 10 tasks. \fourshotfirst{Orange} and \fourshotsecond{green} highlight the best and second-best results per task.
}
\label{tab:cobsat:4shot}
\end{table*}

\subsection{Implementation details}
\label{subsec:implementation_details}
\noindent\textbf{Datasets \& Evaluation.}
Three representative T2I-ICL benchmarks serve as our testbeds: CoBSAT~\cite{cobsat}, Text-to-Image Fast Mini-ImageNet (T2IFMIT)~\cite{tsimpoukelli2021multimodal}, and Dreambench++~\cite{dreambench++}.
We follow their evaluation protocols, where prediction accuracy is reported for all methods on CoBSAT and T2IFMIT, while the normalized scores for concept preservation and prompt following judged by LLMs are reported for all methods on Dreambench++.
Specifically, on CoBSAT, methods receive a task instruction followed by 2 or 4 input image–text pairs (corresponding to the 2-shot and 4-shot settings in our experimental results) and a textual query.
Similarly, on T2IFMIT, the 1-shot and 2-shot settings refer to the number of image–text pairs provided per class.
Following \cite{vlicl}, all experiments on T2IFMIT use three independent random seeds, and we report the mean and standard deviation of performance across all methods.
For Dreambench++, only a text-image pair is provided.
For consistency across experiments, we use the instruction template from~\cite{thinkdiff} and prefix HI’s prompt token with ``Note:''.

\noindent\textbf{Baselines.}
SEED-LLaMA (SL)~\cite{seed_llama}, SEED-X (SX)~\cite{seedx}, Emu~\cite{emu, emu2}, GILL~\cite{gill}, Anole~\cite{anole}, and ThinkDiff~\cite{thinkdiff} are adopted as baselines.
SL-IGC and SX-IGC are the abbreviations for SL and SX fine-tuned on the ImageGen-CoT dataset~\cite{imagegen_cot}, respectively.
In addition, we define a pipeline (denoted as Base) that combines LVLM with FLUX.1-dev~\cite{flux}, the same visual generator adopted in TBDN.
To investigate performance variance introduced by LVLMs, Qwen2-VL-7B-Instruct (Q2)~\cite{qwen2vl}, Qwen2.5-VL-7B-Instruct (Q2.5)~\cite{qwen25vl}, and InternVL3-8B (I3)~\cite{internvl3} are evaluated.

\noindent\textbf{Resources \& Hyperparameters.}
Since our proposed methods are \textbf{training-free}, they require limited computational resources and are easy to reproduce. 
Specifically, the peak memory usage of our methods is under 60 GB, which can be supported by either two consumer-grade GPUs (e.g., RTX 5090) or one professional GPU (e.g., A100).
We set the sampling temperature in LVLM to 0.7, top-p to 0.9, and the number of inference steps in FLUX to 28.
For $\alpha$ in QCD, we set $0.5$ as default.
Additional results and analysis on the effect of $\alpha$ are provided the Appendix.

\subsection{Main results}
\label{sec:main_result}
\noindent\textbf{Results on CoBSAT.}
We evaluate the performance of TBDN and other baselines on 10 tasks in CoBSAT. 
Tables~\ref{tab:cobsat:2shot} and \ref{tab:cobsat:4shot} report the results for the 2-shot and 4-shot settings.
These results across all settings demonstrate that: (i) the simple pipeline, which is denoted as the Base, surprisingly outperforms most unified MLLMs; (ii) Base (Q2) and Base (I3) outperform ThinkDiff, which shares an analogous architecture with the Base, without further modality alignment; (iii) TBDN achieves the best or second-best performance in most cases across different settings.
These results highlight that the paradigm which incorporates the multimodal reasoning capabilities of LVLMs with diffusion models for T2I-ICL is competitive and efficient.
Besides, the effectiveness of TBDN is also demonstrated.

\noindent\textbf{Results on Text-to-Image Fast Mini-ImageNet.}
The main experimental results on Text-to-Image Fast Mini-ImageNet are shown in Table~\ref{tab:t2i_imagenet}, where the I3 identifier is omitted for the Base and TBDN. 
Notably, the Base consistently outperforms other baselines across two settings, indicating a strong advantage in the fast binding T2I-ICL task.
Besides, augmented by our two proposed mechanisms, the mean performance has shown varying degrees of improvement (34.50~$\rightarrow$~39.00, 38.17~$\rightarrow$~39.67) while the standard deviation has decreased accordingly (7.29~$\rightarrow$~2.25, 5.48~$\rightarrow$~2.47).
Comprehensive results of the Base and the TBDN equipped with  various LVLMs are provided in Appendix.

\begin{table}[htbp]
\centering
\begin{tabular}{c|c|c}
\hline
\textbf{Method}         & \textbf{1-shot}       & \textbf{2-shot}       \\ \hline
GILL           & 16.00 ± 2.27 & 15.17 ± 2.72 \\
SL-8B  & 15.00 ± 3.27 & 12.67 ± 1.18 \\
SL-14B & 17.25 ± 2.75 & 16.75 ± 1.75 \\
Emu1       & 31.50 ± 1.87 & 22.83 ± 2.72 \\
Emu2       & 24.33 ± 3.30 & 30.67 ± 1.31 \\
Anole-7B       & 11.00 ± 2.86 & 7.00 ± 0.71  \\ \hline


Base      & 34.50 ± 7.29 & 38.17 ± 5.48 \\
+ HI           & 36.50 ± 1.53 & 38.00 ± 2.18 \\
TBDN    & \textbf{39.00 ± 2.25} & \textbf{39.67 ± 2.47}
\\ \hline
\end{tabular}
\caption{Comparison of accuracy on T2IFMIT.}
\label{tab:t2i_imagenet}
\end{table}

\noindent\textbf{Results on Dreambench++.}
Table~\ref{tab:dreambench++} shows the brief comparison results on Dreambench++.
Compared to the two aforementioned benchmarks, Dreambench++ tends to measure the ability of methods to capture and processing the visual detail, where TBDN achieve promising scores in prompt following (PF) but underperforms in concept preservation (CP).
This result is expected because the ability for CP is inherent in TBDN’s visual generator, which is fixed, and the essence of PF aligns with the design of HI and QCD, indicating directions for future improvements. 
More comprehensive results are provided in the Appendix.
\begin{table}[htbp]
\resizebox{0.49\textwidth}{!}{
\begin{tabular}{c|c|c|c}
\hline
\textbf{\begin{tabular}[c]{@{}c@{}}Dream \\ bench++ \end{tabular}} & \textbf{\begin{tabular}[c]{@{}c@{}}Concept \\ Preservation \end{tabular}} & \textbf{\begin{tabular}[c]{@{}c@{}}Prompt \\ Following \end{tabular}} & \textbf{CP·PF~$\uparrow$} \\ \hline

SL              & .358                                                                          & .218                                                                      & .078           \\
SL-IGC          & .325                                                                          & .310                                                                      & .101           \\
SX              & \textbf{.559}                                                                          & .337                                                                      & .188           \\
SX-IGC          & \underline{.458}                                                                          & \textbf{.881}                                                                      & \textbf{.403}           \\ \hline
TBDN~(Q2)        & .442                                                                          & \underline{.778}                                                                      & \underline{.344}           \\ \hline
\end{tabular}
}
\caption{Comparison of two LLM-judged scores.}
\label{tab:dreambench++}
\end{table}


\noindent\textbf{Ablation of HI and QCD.}
Our ablation results across LVLMs are summarized in Tables~\ref{tab:abla:internvl3},~\ref{tab:abla:qwen2}, and~\ref{tab:abla:qwen2.5}.
Across settings, HI and QCD consistently enhance the Base method, securing the best and second-best performances in most cases, which demonstrates a direct outcome of their complementary strengths.
\begin{table*}[]
\centering
\resizebox{\textwidth}{!}{
\begin{tabular}{c|c|ccccc|ccccc|l}
\hline
\multirow{2}{*}{\textbf{Method}}               & \multirow{2}{*}{\textbf{Shot}} & \multicolumn{5}{c|}{\textbf{Object-Inference Task}}        & \multicolumn{5}{c|}{\textbf{Attribute-Inference Task}}          & \multirow{2}{*}{\textbf{Avg. acc.}$\uparrow$} \\ \cline{3-12}
                                      &                       & Color-I & Bkg.-I & Style-I & Action-I & Texture-I & Color-II & Bkg.-II & Style-II & Action-II & Texture-II &                       \\ \hline
\multirow{2}{*}{Base~(I3)}         & 2                     & .826   & .760   & .317   & .711    & .432     & .385    & .866   & .442    & .701     & .429      & .586                 \\
                                      & 4                     & .881   & .908  & .401   & .824    & .573     & .659    & .936   & .522    & .837     & .588      & .712                 \\ \hline
\multirow{2}{*}{+ HI}        & 2                     & .813   & .739  & .225   & .627    & .385     & .285    & .878   & .334    & \twoshotfirst{.808}     & .362      & .545($\downarrow$7.0\%)                 \\
                                      & 4                     & .846   & \fourshotsecond{.924}  & .291   & .741    & .499     & .397    & .964   & .403    & \fourshotfirst{.882}     & .497      & .644($\downarrow$9.6\%)                  \\ \hline
\multirow{2}{*}{+ QCD}                & 2                     & \twoshotsecond{.913}   & \twoshotsecond{.711}  & \twoshotfirst{.421}   & \twoshotfirst{.850}     & \twoshotsecond{.513}     & \twoshotsecond{.540}     & \twoshotsecond{.924}   & \twoshotfirst{.522}    & .637     & \twoshotsecond{.518}      & \twoshotsecond{.654}($\uparrow$11.6\%)                 \\
                                      & 4                     & \fourshotsecond{.908}   & .921  & \fourshotfirst{.554}   & \fourshotfirst{.907}    & \fourshotfirst{.638}     & \fourshotsecond{.726}    & \fourshotsecond{.981}   & \fourshotfirst{.566}    & .807     & \fourshotsecond{.624}      & \fourshotsecond{.763}($\uparrow$7.2\%)                 \\ \hline
\multirow{2}{*}{+ QCD \& HI} & 2                     & \twoshotfirst{.927}   & \twoshotfirst{.808}  & \twoshotsecond{.383}   & \twoshotsecond{.808}    & \twoshotfirst{.520}      & \twoshotfirst{.587}    & \twoshotfirst{.956}   & \twoshotsecond{.486}   & \twoshotsecond{.789}     & \twoshotfirst{.568}      & \twoshotfirst{.683}($\uparrow$16.6\%)                 \\
                                      & 4                     & \fourshotfirst{.927}   & \fourshotfirst{.951}  & \fourshotsecond{.490}    & \fourshotsecond{.875}    & \fourshotsecond{.627}     & \fourshotfirst{.783}    & \fourshotfirst{.983}   & \fourshotsecond{.547}    & \fourshotsecond{.868}     & \fourshotfirst{.636}      & \fourshotfirst{.768}($\uparrow$7.9\%)                \\ \hline
\end{tabular}
}
\caption{
Ablation of HI and QCD with InternVL3 on CoBSAT benchmark.
\twoshotfirst{Red} and \twoshotsecond{blue} highlight the best and second-best 2-shot results per task, while \fourshotfirst{Orange} and \fourshotsecond{green} highlight the best and second-best 4-shot results.
}
\label{tab:abla:internvl3}
\end{table*}
For Qwen2-VL and InternVL3, HI and QCD exhibit distinct non-overlapping advantages. 
HI excels in Bkg.-I and Action-II tasks while QCD dominates other subtasks.
This division of strengths aligns with the complementary concept we introduced earlier.
HI and QCD mirror two halves of a \textit{Möbius Band}-like closed-loop constraint in that neither is complete on its own while their integration delivers a continuous and comprehensive solution.

\noindent\textbf{Instruction comparison.}
As noted in Sec.~\ref{sec:method:qcd}, for a given dataset, input differences across methods stem from $X_{\text{ins}}$ variations, which induce distinct costs in memory footprint and API call quotas. 
To evaluate the effectiveness and efficiency of HI, we compare it with 4 instructions: (1) CB-Ins; (2) CoT-Ins; (3) TD-Ins; (4) TD-Ins++.
CB-Ins and CoT-Ins are from~\cite{cobsat}, and TD-Ins from~\cite{thinkdiff}.
TD-Ins++ is TD-Ins suffixed with ``Let's think it step by step''.
We report average accuracy, performance improvement ($\delta$), and average instruction token length across two settings.

\begin{table}[htbp]
\centering
\begin{tabular}{c|c|c|c|c}
\hline
\textbf{Qwen2-VL} & \textbf{Shot} & \textbf{Acc.} & $\mathbf{\delta}$ & \textbf{Len.}  \\ \hline
\multirow{2}{*}{CB-Ins} & 2             & .447      & --         & \multirow{2}{*}{42}      \\ \cline{2-4} 
                          & 4             & .353     & --         &                   \\ \hline
\multirow{2}{*}{CoT-Ins}      & 2             & .533   &.086           & 2850         \\ \cline{2-5} 
                          & 4             & .642     &.289         & 5521           \\ \hline
\multirow{2}{*}{TD-Ins}     & 2             & .537       &.090      & \multirow{2}{*}{48}  \\ \cline{2-4} 
                          & 4             & .614       &.261       &                   \\ \hline  
\multirow{2}{*}{TD-Ins++}     & 2             & .561       &.114      & \multirow{2}{*}{55}  \\ \cline{2-4} 
                          & 4             & --       &--       &                   \\ \hline                            
\multirow{2}{*}{HI} & 2             & \textbf{.601}   & \textbf{.154}   & \multirow{2}{*}{82}  \\ \cline{2-4} 
                          & 4             & \textbf{.673}  & \textbf{.320}   &                  \\ \hline                          
\end{tabular}
\caption{The average of accuracy and token length about different instructions in CoBSAT.}
\label{tab: ins_comparision}
\end{table}
Table~\ref{tab: ins_comparision} shows CoT-Ins, TD-Ins, TD-Ins++, and HI all yield performance gains relative to CB-Ins, with HI achieving the most substantial improvement. 
Notably, CB-Ins, TD-Ins, TD-Ins++, and HI have comparable token lengths, while CoT-Ins exhibits an extremely large token count. 
This is because CoT-Ins requires inserting in-context samples to construct reasoning chains, drastically raising token usage.
As the number of shots increases, this token overhead escalates further and may even exceed the context window limit of some LVLMs (e.g., InternVL3), rendering CoT-Ins impractical for large-shot scenarios.
CB-Ins is a prompt variant that explicitly states the object–attribute relationship in the instruction. 
For instance, for the Color-I task in CoBSAT, CB-Ins must be modified to incorporate the ``color signal": ``\textit{Please identify the main common object in the images, and generate another image of this object in the requested color.}" 
While CB-Ins is shorter than HI, it suffers from poor generalization due to the need for task-specific rewriting. 
Moreover, LVLMs often fail to correctly interpret CB-Ins and even produce irrelevant responses in our empirical practice.
In contrast, HI eliminates the need for task-specific redesign across sub-tasks, simultaneously delivering superior performance efficiently and striking a balance between effectiveness and generalizability.
Additionally, as mentioned in Sec.~\ref{subsec:implementation_details}, HI adopts the instruction template from TD-Ins.
Compared with TD-Ins and TD-Ins++, HI achieves substantial improvement with minimal extra token cost.

\noindent\textbf{HI Variant Comparison.}
\begin{figure}[htbp]
    \centering
    \includegraphics[width=0.42\textwidth]{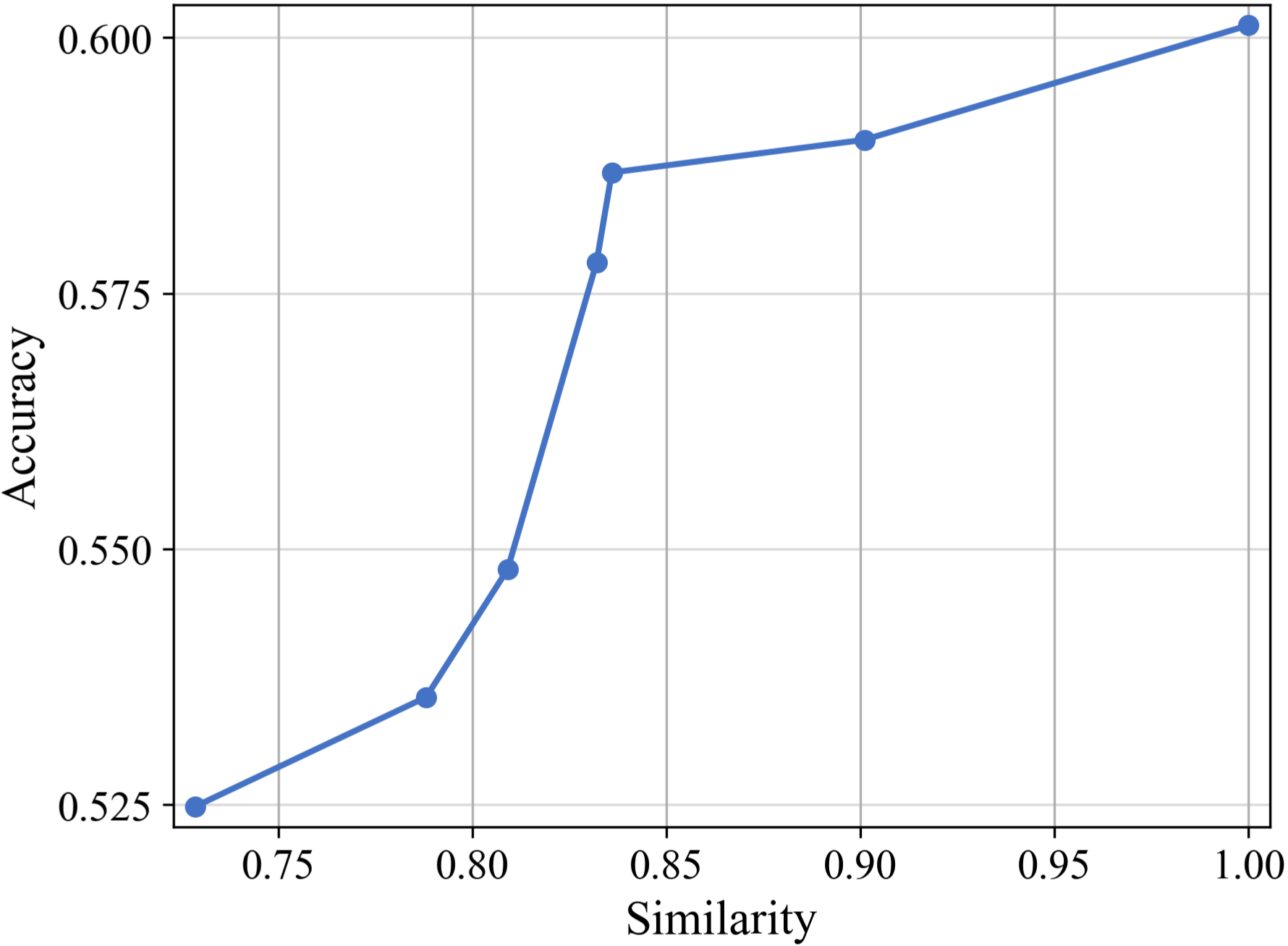}
    \caption{Average accuracy on CoBSAT and sentences similarity between canonical HI prompt and its variants.}
    \label{fig: sentences_similarity}
\end{figure}
A natural concern in HI design is whether prompt variations could lead to performance inconsistency.
To address this, we propose and validate a reproducible recipe that incorporates a canonical prompt and a semantic similarity threshold to facilitate reliable prompt adaptation.
To instantiate this recipe, we conduct a three-stage implementation as follows.
We manually construct an initial prompt following the aforementioned two principles (see Sec.~\ref{sec:method:hint}).
We then feed this initial prompt into GPT-4o~\cite{gpt4} to generate prompt variants, which we subsequently evaluate on CoBSAT to select the candidate with best performing as the canonical prompt of HI.
Using sentence-level embeddings generated from BGE-M3~\cite{bgem3}, we compute dot product similarities between each variant and the canonical prompt.
As illustrated in Fig.~\ref{fig: sentences_similarity}, variants with a similarity $\geq$ 0.80 consistently achieve an accuracy $\geq$ 0.54, outperforming the TD-Ins (see Table~\ref{tab: ins_comparision}).
We report the list of prompt variants and their corresponding accuracies in Table~\ref{tab: ins_variants} (Appendix).

\section{Conclusion}

In this paper, we propose TBDN, a training free framework that integrates two complementary enhancement mechanisms in T2I-ICL. 
By explicitly addressing two recurrent failure patterns identified in our study, compliance failure and prior dominated hallucination, TBDN improves rule following and reduces prior override during in context generation. 
Across CoBSAT, T2I Fast Mini-ImageNet, and Dreambench++, TBDN consistently achieves strong performance and generalization under diverse settings. 
Extensive ablation studies further confirm its robustness across model backbones and hyperparameter choices.

\section{Limitations}
\label{sec:limitations}
Despite the effectiveness of TBDN, it has several limitations.
First, since LVLMs cannot generate images directly, TBDN relies on instructions to guide LVLMs in producing textual descriptions for the T2I generator.
This indirect design risks semantic gaps between text and images, unlike MLLMs that enable end-to-end multimodal-to-image generation.
Besides, TBDN is less suitable for multimodal image composing tasks (e.g., ~\cite{dreambench++}), as our design prioritizes context-aware query reasoning over modeling of fine-grained visual details in reference images.
Finally, the generalization of our designs for HI and QCD  to MLLMs remains under-explored.
Future work will address these gaps by exploring end-to-end multimodal generation paradigms, enhancing fine-grained visual alignment, and extending our designs to MLLMs.




\bibliography{custom}

\appendix

\section{Implementation Details}
\begin{figure*}[htbp]
\centering
\includegraphics[width=\textwidth]{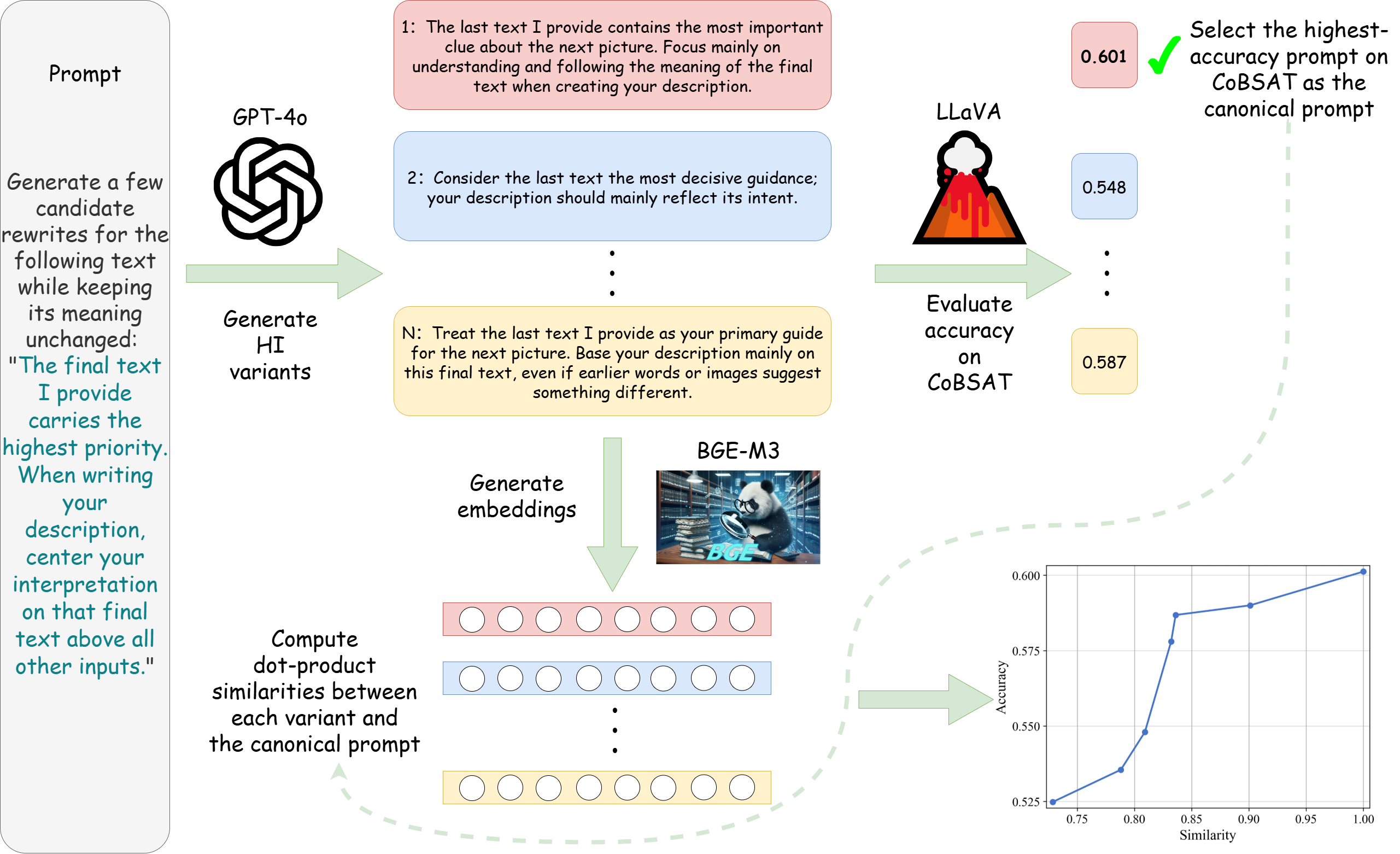}
  \caption{Overview of the pipeline for selecting the canonical Hint Instruction and measuring semantic similarity.}
  \label{fig:hint_prompt_pipeline}
\end{figure*}

Fig.~\ref{fig:hint_prompt_pipeline} illustrates the details of the HI Variant Comparison.
The list of prompt variants and their corresponding accuracies is summarized in Table~\ref{tab: ins_variants}.

In Table~\ref{tab: ins_comparision}, we compare the performance of different instruction variants on the CoBSAT. 
CB-Ins and CoT-Ins are from the CoBSAT paper, while TD-Ins is adopted from the ThinkDiff paper. 
Both HI and TD-Ins++ are derived from TD-Ins: HI uses TD-Ins as the basic instruction, and TD-Ins++ further augments TD-Ins by adding a naive CoT prompt.
Below is the list of detailed instructions:
\begin{itemize}
    \item \textbf{CB-Ins (For Color-I task)}: \textit{Please identify the common main object in the images, and describe the next image to be generated based on the sequence below. Your description of image should contain the description of the common main object and the requested color. }
    \item \textbf{CoT-Ins}: \textit{We provide a few examples, each of which is an input-output pair where the output is a description of the image associated with the input.\\
    (Multimodal context)
    \\
    Based on the examples, the task is to predict the next image description. Before predicting the next image, let's think step by step and analyze what the relationship between the text input and image output in each example is first. 
    \\
    (Model's response)
    \\
    Based on the analysis, please describe what the next image should be look like given the request.}
    \item \textbf{TD-Ins}: \textit{I give you several words and pictures. First, please analyse what the next picture is. Then give me a detailed diffusion prompt to describe the next picture. Please only provide me the detailed prompt and start the answer with `Create an image'}.
    \item \textbf{TD-Ins++}: \textit{I give you several words and pictures. First, please analyse what the next picture is. Then give me a detailed diffusion prompt to describe the next picture. Please only provide me the detailed prompt and start the answer with `Create an image'. \textbf{Let's think step by step}}.
    \item \textbf{HI}: \textit{I give you several words and pictures. First, please analyse what the next picture is. 
    Then give me a detailed diffusion prompt to describe the next picture. Please only provide me the detailed prompt and start the answer with `Create an image'. 
    \textbf{Note: The last text I provide contains the most important clue about the next picture. 
    Focus mainly on understanding and following the meaning of the final text when creating your description.}}
\end{itemize}

\begin{table*}[htbp]
\resizebox{\textwidth}{!}{
\begin{tabular}{c|c|c}
\hline
\textbf{Hint Instructions}                                                                                                                                                                                             & \textbf{Similarity} & \textbf{Avg. Acc.} \\ \hline
\begin{tabular}[c]{@{}c@{}}The last text I provide contains the most important clue about the next picture. \\ Focus mainly on understanding and following the meaning of the final text when creating your description.\end{tabular}  & 1.000      & \textbf{0.601}            \\ \hline
\begin{tabular}[c]{@{}c@{}}Consider the last text the most decisive guidance; \\ your description should mainly reflect its intent.\end{tabular}                                                                                       & 0.688      & 0.548            \\ \hline
\begin{tabular}[c]{@{}c@{}}The concluding text is the key signal—base your description chiefly \\ on it, giving it precedence over the rest of the context.\end{tabular} 
                                                                                                          & 0.684      & 0.525            \\ \hline
\begin{tabular}[c]{@{}c@{}}Treat the last text as the decisive guide, and make sure your description \\ 
is driven mainly by its intent. \end{tabular}                                                                                                   
                                                                                                                             & 0.681      & 0.536            \\ \hline
\begin{tabular}[c]{@{}c@{}}Treat the last text I provide as your primary guide for the next picture. \\ Base your description mainly on this final text, even if earlier words or images suggest something different.\end{tabular}     & 0.788      & 0.587            \\ \hline
\end{tabular}
}
\caption{Detail hint instructions and corresponding sentences similarity and average accuracy.}
\label{tab: ins_variants}
\end{table*}

\section{More quantitative results}
\label{sec:more_quantitative_results}

\subsection{More ablation studies of HI and QCD across different LVLMs}
\label{subsecion:ablation_HI_QCD_LVLMs}
Table~\ref{tab:abla:qwen2} and \ref{tab:abla:qwen2.5} present the ablation evaluation of HI and QCD associated with Qwen2-VL and Qwen2.5VL.
Taking the result of 2-shot setting as an example, HI increases the performance of Qwen2-VL from 0.537 to 0.601 (+11.9\%) and that of Qwen2.5-VL from 0.312 to 0.357 (+14.4\%). 
It suggests that, beyond explicit instructions and prompt-based methods such as chain-of-thought (CoT), incorporating prior knowledge as a hint condition can further enhance the model’s adherence to contextual queries, reduce subject-related errors, and thereby improve text-to-image generation accuracy.
Generally, compared with HI, QCD yields even larger performance gains.
For Qwen2-VL, it achieves 0.638 (+18.8\%) and 0.745 (+21.3\%) under the 2-shot and 4-shot settings, respectively. 
For Qwen2.5-VL, the improvements are even more substantial, with scores increasing from 0.312 to 0.554 (+77.6\%) in the 2-shot setting and from 0.394 to 0.634 (+60.9\%) in the 4-shot setting. 
These results indicate that QCD can mitigate the language bias of LVLMs induced by in-context information, enabling them to describe the target image more accurately and thus generate more faithful images.

\begin{table*}[htbp]
\resizebox{\textwidth}{!}{
\begin{tabular}{c|c|ccccc|ccccc|l}
\hline
\multirow{2}{*}{\textbf{Method}}               & \multirow{2}{*}{\textbf{Shot}} & \multicolumn{5}{c|}{\textbf{Object-Inference Task}}        & \multicolumn{5}{c|}{\textbf{Attribute-Inference Task}}          & \multirow{2}{*}{\textbf{Avg. acc.}$\uparrow$} \\ \cline{3-12}
                                      &                       & Color-I & Bkg.-I & Style-I & Action-I & Texture-I & Color-II & Bkg.-II & Style-II & Action-II & Texture-II &                                 \\ \hline
\multirow{2}{*}{Base~(Q2)}    & 2                     & .766    & .719   & .299    & .498     & .426      & .501     & .769    & .338     & .686      & .366       & .537                            \\
                                      & 4                     & .815    & .730   & .346    & .466     & .504      & .652     & .920     & .443     & \fourshotsecond{.812}      & .449       & .614                            \\ \hline
\multirow{2}{*}{+ HI}        & 2                     & .811    & \twoshotsecond{.797}   & .323    & .512     & .487      & \twoshotsecond{.664}     & .847    & .416     & \twoshotfirst{.722}      & .433       & .601($\uparrow$11.9\%)                    \\
                                      & 4                     & .824    & .766   & .385    & .555     & .554      & \fourshotfirst{.832}     & .958    & .506     & \fourshotfirst{.850}      & .497       & .673($\uparrow$9.6\%)                     \\ \hline
\multirow{2}{*}{+ QCD}                & 2                     & \twoshotsecond{.866}    & .743   & \twoshotsecond{.442}    & \twoshotsecond{.721}     & \twoshotsecond{.529}      & .621     & \twoshotsecond{.875}    & \twoshotsecond{.455}     & .642      & \twoshotsecond{.489}      & \twoshotsecond{.638}($\uparrow$18.8\%)                    \\
                                      & 4                     & \fourshotfirst{.946}    & \fourshotsecond{.904}   & \fourshotsecond{.561}    & \fourshotsecond{.786}     & \fourshotsecond{.616}      & \fourshotsecond{.800}     & \fourshotsecond{.963}    & \fourshotfirst{.552}     & .791      & \fourshotsecond{.533}       & \fourshotsecond{.745}($\uparrow$21.3\%)                    \\ \hline
\multirow{2}{*}{TBDN} & 2                     & \twoshotfirst{.909}    & \twoshotfirst{.879}   & \twoshotfirst{.500}    & \twoshotfirst{.760}     & \twoshotfirst{.556}      & \twoshotfirst{.724}     & \twoshotfirst{.905}    & \twoshotfirst{.487}     & \twoshotsecond{.683}      & \twoshotfirst{.531}       & \twoshotfirst{.693}($\uparrow$29.1\%)           \\
                                      & 4                     & \fourshotsecond{.920}    & \fourshotfirst{.948}   & \fourshotfirst{.582}    & \fourshotfirst{.822}     & \fourshotfirst{.633}      & \fourshotfirst{.832}    & \fourshotfirst{.980}    & \fourshotsecond{.550}     & .788      & \fourshotfirst{.593}       & \fourshotfirst{.767}($\uparrow$24.9\%)           \\ \hline
\end{tabular}
}
\caption{Ablation of HI and QCD on Qwen2-VL on CoBSAT benchmark. \twoshotfirst{Red} and \twoshotsecond{blue} highlight the best and second-best 2-shot results per task, while \fourshotfirst{Orange} and \fourshotsecond{green} highlight the best and second-best 4-shot results.}
\label{tab:abla:qwen2}
\end{table*}
\begin{table*}[htbp]
\resizebox{\textwidth}{!}{
\begin{tabular}{c|c|ccccc|ccccc|l}
\hline
\multirow{2}{*}{\textbf{Method}}               & \multirow{2}{*}{\textbf{Shot}} & \multicolumn{5}{c|}{\textbf{Object-Inference Task}}        & \multicolumn{5}{c|}{\textbf{Attribute-Inference Task}}          & \multirow{2}{*}{\textbf{Avg. acc.}$\uparrow$} \\ \cline{3-12}
                                      &                       & Color-I & Bkg.-I & Style-I & Action-I & Texture-I & Color-II & Bkg.-II & Style-II & Action-II & Texture-II &                       \\ \hline
\multirow{2}{*}{Base~(Q2.5)}           & 2                     & .354   & .407  & .127   & .392    & .223     & .231    & .442   & .213    & .522     & .217      & .312                 \\
                                      & 4                     & .498   & .558  & .149   & .446    & .255     & .369    & .493   & .265    & .620      & .296      & .394                 \\ \hline
\multirow{2}{*}{+ HI}        & 2                     & .426   & .481  & .153   & .439    & .237     & .294    & .509   & .236    & .526     & .278      & .357($\uparrow$14.4\%)                 \\
                                      & 4                     & .619   & .590   & .201   & .521    & .327     & .536    & .628   & .345    & .686     & .393      & .484($\uparrow$22.8\%)                 \\ \hline
\multirow{2}{*}{+ QCD}                & 2                     & \twoshotsecond{.770}    & \twoshotsecond{.526}  & \twoshotsecond{.295}   & \twoshotfirst{.650}     & \twoshotsecond{.424}     & \twoshotfirst{.555}    & \twoshotfirst{.840}    & \twoshotsecond{.399}    & \twoshotfirst{.664}     & \twoshotsecond{.425}      & \twoshotsecond{.554}($\uparrow$77.6\%)                 \\
                                      & 4                     & \fourshotsecond{.811}   & \fourshotsecond{.816}  & \fourshotsecond{.322}   & \fourshotsecond{.679}    & \fourshotsecond{.452}     & \fourshotsecond{.647}    & \fourshotsecond{.945}   & \fourshotsecond{.436}    & \fourshotfirst{.745}     & \fourshotsecond{.496}      & \fourshotsecond{.634}($\uparrow$60.9\%)                 \\ \hline
\multirow{2}{*}{+ QCD \& HI} & 2                     & \twoshotfirst{.843}   & \twoshotfirst{.594}  & \twoshotfirst{.347}   & \twoshotsecond{.624}    & \twoshotfirst{.459}     & \twoshotsecond{.534}    & \twoshotsecond{.794}   & \twoshotfirst{.409}    & \twoshotsecond{.589}     & \twoshotfirst{.441}      & \twoshotfirst{.563}($\uparrow$80.4\%)                \\
                                      & 4                     & \fourshotfirst{.868}   & \fourshotfirst{.832}  & \fourshotfirst{.392}   & \fourshotfirst{.752}    & \fourshotfirst{.531}     & \fourshotfirst{.675}    & \fourshotfirst{.960}    & \fourshotfirst{.457}    & \fourshotsecond{.740}      & \fourshotfirst{.517}      & \fourshotfirst{.672}($\uparrow$70.6\%)                 \\ \hline
\end{tabular}
}
\caption{
Ablation of HI and QCD with Qwen2.5-VL on CoBSAT benchmark.
\twoshotfirst{Red} and \twoshotsecond{blue} highlight the best and second-best 2-shot results per task, while \fourshotfirst{Orange} and \fourshotsecond{green} highlight the best and second-best 4-shot results.
}
\label{tab:abla:qwen2.5}
\end{table*}

\subsection{Effect of $\alpha$ in QCD.}
Table \ref{tab:abla:alpha} summarizes the performance of LVLMs enhanced by QCD with different $\alpha$ values on CoBSAT.
Qwen2-VL shows a performance drop when $\alpha$ increases from 0.5 to 1.0, while Qwen2.5-VL achieves peak accuracy at $\alpha = 0.75$.
InternVL3 exhibits minimal performance fluctuation over 4 $\alpha$ settings.
The optimal performance across three LVLMs is consistently attained at intermediate $\alpha$ values.
Based on these observations, we adopt $\alpha = 0.5$ as the default setting for TBDN.
\begin{table}[tbhp]
\centering
\begin{tabular}{c|ccc}
\hline
$\mathbf{\alpha}$ & \textbf{Base~(Q2)}      & \textbf{Base~(Q2.5)}    & \textbf{Base~(I3)}      \\ \hline
0.25    & .632          & .478          & .643          \\
0.5     & \textbf{.638} & .555          & \textbf{.655} \\
0.75    & .614          & \textbf{.575} & .653          \\
1.0     & .552          & .571          & .647          \\ \hline
\end{tabular}
\caption{Average accuracy of LVLMs with different $\alpha$.}
\label{tab:abla:alpha}
\end{table}

\subsection{Text version results on CoBSAT}
\label{subsec:text_cobsat}
We present a text-version results on the CoBSAT benchmark, as shown in Table \ref{tab:cobsat:text}.
Similar to the image-generation setting defined in CoBSAT, our method delivers consistent improvements across three different LVLMs under both 2-shot and 4-shot configurations in the text-only setting.
Meanwhile, our method achieves the best and second-best results, outperforming the strong baseline SX-IGC as well as the proprietary model Gemini~\cite{gemini}.
These results suggest that the high accuracy observed in image-generation mode is grounded in accurate textual descriptions.

In addition, compared with ThinkDiff and several MLLMs, our method offers stronger interpretability.
Consistent with the findings in~\cite{imagegen_cot}, however, accuracy in the image-generation mode remains lower than that in text-generation mode, indicating that reliably translating precise textual descriptions into equally accurate images remains a non-trivial challenge.

\begin{table*}[htpb]
\centering
\resizebox{\textwidth}{!}{
\begin{tabular}{ccccccccccccc}
\hline
\multicolumn{1}{c|}{\multirow{2}{*}{\textbf{Method}}}               & \multicolumn{1}{l|}{\multirow{2}{*}{\textbf{Shot}}} & \multicolumn{5}{c|}{\textbf{Object-Inference Task}}                             & \multicolumn{5}{c|}{\textbf{Attribute-Inference Task}}                               & \multicolumn{1}{c}{\multirow{2}{*}{\textbf{Avg. acc.}$\uparrow$}} \\ \cline{3-12}
\multicolumn{1}{c|}{}                                      & \multicolumn{1}{l|}{}                      & Color-I & Bkg.-I & Style-I & Action-I & \multicolumn{1}{c|}{Texture-I} & Color-II & Bkg.-II & Style-II & Action-II & \multicolumn{1}{c|}{Texture-II} & \multicolumn{1}{l}{}                           \\ \hline
\multicolumn{13}{l}{\textit{Qwen2-VL}}                                                                                                                                                                                                                                                                          \\ \hline
\multicolumn{1}{c|}{\multirow{2}{*}{Base}}                 & \multicolumn{1}{c|}{2}                     & .810    & .757   & .357    & .561     & \multicolumn{1}{c|}{.606}      & .517     & .762    & .440     & .737      & \multicolumn{1}{c|}{.474}       & \multicolumn{1}{l}{.602}                                           \\
\multicolumn{1}{c|}{}                                      & \multicolumn{1}{c|}{4}                     & .850    & .764   & .447    & .540     & \multicolumn{1}{c|}{.711}      & .683     & .919    & .578     & .848      & \multicolumn{1}{c|}{.635}       & \multicolumn{1}{l}{.698}                                           \\ \hline
\multicolumn{1}{c|}{\multirow{2}{*}{+ HI}}                 & \multicolumn{1}{c|}{2}                     & .848    & \twoshotsecond{.851}   & .430    & .618     & \multicolumn{1}{c|}{.668}      & .697     & .834    & .541     & .778      & \multicolumn{1}{c|}{.550}       & \multicolumn{1}{l}{.682($\uparrow$13.3\%)}                                           \\
\multicolumn{1}{c|}{}                                      & \multicolumn{1}{c|}{4}                     & .862    & .812   & .512    & .657     & \multicolumn{1}{c|}{.779}      & .871     & .961    & .630     & .885      & \multicolumn{1}{c|}{.748}       & \multicolumn{1}{l}{.772($\uparrow$10.6\%)}                                           \\ \hline
\multicolumn{1}{c|}{\multirow{2}{*}{+ QCD}}                & \multicolumn{1}{c|}{2}                     & .910    & .764   & .583    & .829     & \multicolumn{1}{c|}{\twoshotsecond{.739}}      & .655     & .872    & .587     & .699      & \multicolumn{1}{c|}{.637}       & \multicolumn{1}{l}{.728($\uparrow$20.9\%)}                                           \\
\multicolumn{1}{c|}{}                                      & \multicolumn{1}{c|}{4}                     & .970    & .916   & .725    & .888     & \multicolumn{1}{c|}{\fourshotsecond{.885}}      & .844     & .972    & .684     & .833      & \multicolumn{1}{c|}{.747}       & \multicolumn{1}{l}{.846($\uparrow$21.2\%)}                                           \\ \hline
\multicolumn{1}{c|}{\multirow{2}{*}{+ QCD \& HI}}          & \multicolumn{1}{c|}{2}                     & .933    & \twoshotfirst{.898}   & \twoshotsecond{.674}    & .862     & \multicolumn{1}{c|}{\twoshotfirst{.789}}      & \twoshotfirst{.784}     & .896    & .640     & .725      & \multicolumn{1}{c|}{\twoshotsecond{.706}}       & \multicolumn{1}{l}{.791($\uparrow$31.4\%)}                                           \\
\multicolumn{1}{c|}{}                                      & \multicolumn{1}{c|}{4}                     & \fourshotsecond{.949}    & \fourshotsecond{.965}   & \fourshotfirst{.754}    & .922     & \multicolumn{1}{c|}{\fourshotfirst{.891}}      & \fourshotfirst{.922}     & .980    & \fourshotsecond{.719}     & .866      & \multicolumn{1}{c|}{\fourshotsecond{.815}}       & \multicolumn{1}{l}{.878($\uparrow$25.8\%)}                                           \\ \hline
\multicolumn{13}{l}{\textit{Qwen2.5-VL}}                                                                                                                                                                                                                                                                        \\ \hline
\multicolumn{1}{c|}{\multirow{2}{*}{Base}}                 & \multicolumn{1}{c|}{2}                     & .363    & .439   & .215    & .483     & \multicolumn{1}{c|}{.316}      & .239     & .440    & .270     & .543      & \multicolumn{1}{c|}{.275}       & \multicolumn{1}{l}{.358}                                           \\
\multicolumn{1}{c|}{}                                      & \multicolumn{1}{c|}{4}                     & .517    & .586   & .222    & .532     & \multicolumn{1}{c|}{.346}      & .386     & .494    & .332     & .650      & \multicolumn{1}{c|}{.396}       & \multicolumn{1}{l}{.446}                                           \\ \hline
\multicolumn{1}{c|}{\multirow{2}{*}{+ HI}}                 & \multicolumn{1}{c|}{2}                     & .440    & .517   & .272    & .524     & \multicolumn{1}{c|}{.323}      & .315     & .502    & .294     & .555      & \multicolumn{1}{c|}{.365}       & \multicolumn{1}{l}{.411($\uparrow$14.8\%)}                                           \\
\multicolumn{1}{c|}{}                                      & \multicolumn{1}{c|}{4}                     & .647    & .628   & .289    & .626     & \multicolumn{1}{c|}{.473}      & .558     & .630    & .431     & .702      & \multicolumn{1}{c|}{.508}       & \multicolumn{1}{l}{.549($\uparrow$23.1\%)}                                           \\ \hline
\multicolumn{1}{c|}{\multirow{2}{*}{+ QCD}}                & \multicolumn{1}{c|}{2}                     & .801    & .531   & .430    & .785     & \multicolumn{1}{c|}{.624}      & .550     & .841    & .496     & .684      & \multicolumn{1}{c|}{.558}       & \multicolumn{1}{l}{.630($\uparrow$76.0\%)}                                           \\
\multicolumn{1}{c|}{}                                      & \multicolumn{1}{c|}{4}                     & .834    & .832   & .452    & .812     & \multicolumn{1}{c|}{.666}      & .670     & .951    & .559     & .813      & \multicolumn{1}{c|}{.650}       & \multicolumn{1}{l}{.724($\uparrow$62.3\%)}                                           \\ \hline
\multicolumn{1}{c|}{\multirow{2}{*}{+ QCD \& HI}}          & \multicolumn{1}{c|}{2}                     & .878    & .615   & .530    & .774     & \multicolumn{1}{c|}{.628}      & .577     & .806    & .507     & .620      & \multicolumn{1}{c|}{.603}       & \multicolumn{1}{l}{.654($\uparrow$82.7\%)}                                           \\
\multicolumn{1}{c|}{}                                      & \multicolumn{1}{c|}{4}                     & .915    & .862   & .605    & .892     & \multicolumn{1}{c|}{.752}      & .703     & .960    & .576     & .785      & \multicolumn{1}{c|}{.688}       & \multicolumn{1}{l}{.774($\uparrow$73.5\%)}                                           \\ \hline
\multicolumn{13}{l}{\textit{InternVL3}}                                                                                                                                                                                                                                                                         \\ \hline
\multicolumn{1}{c|}{\multirow{2}{*}{Base}}                 & \multicolumn{1}{c|}{2}                     & .867    & .775   & .418    & .832     & \multicolumn{1}{c|}{.601}      & .401     & .869    & .543     & .750      & \multicolumn{1}{c|}{.536}       & \multicolumn{1}{l}{.659}                                           \\
\multicolumn{1}{c|}{}                                      & \multicolumn{1}{c|}{4}                     & .917    & .930   & .510    & .944     & \multicolumn{1}{c|}{.810}      & .682     & .947    & .644     & .866      & \multicolumn{1}{c|}{.770}       & \multicolumn{1}{l}{.802}                                           \\ \hline
\multicolumn{1}{c|}{\multirow{2}{*}{+ HI}}                 & \multicolumn{1}{c|}{2}                     & .857    & .780   & .343    & .751     & \multicolumn{1}{c|}{.544}      & .309     & .884    & .449     & \twoshotfirst{.848}      & \multicolumn{1}{c|}{.473}       & \multicolumn{1}{l}{.624($\downarrow$5.3\%)}                                           \\
\multicolumn{1}{c|}{}                                      & \multicolumn{1}{c|}{4}                     & .870    & .945   & .408    & .859     & \multicolumn{1}{c|}{.720}      & .427     & .966    & .525     & \fourshotsecond{.923}      & \multicolumn{1}{c|}{.679}       & \multicolumn{1}{l}{.732($\downarrow$8.7\%)}                                           \\ \hline
\multicolumn{1}{c|}{\multirow{2}{*}{+ QCD}}                & \multicolumn{1}{c|}{2}                     & .949    & .723   & .617    & \twoshotsecond{.956}     & \multicolumn{1}{c|}{.709}      & .587     & .927    & \twoshotsecond{.662}     & .665      & \multicolumn{1}{c|}{.674}       & \multicolumn{1}{l}{.747($\uparrow$13.4\%)}                                           \\
\multicolumn{1}{c|}{}                                      & \multicolumn{1}{c|}{4}                     & .947    & .940   & \fourshotsecond{.748}    & \fourshotfirst{.985}     & \multicolumn{1}{c|}{.875}      & .757     & \fourshotsecond{.984}    & \fourshotfirst{.726}     & .849      & \multicolumn{1}{c|}{.814}       & \multicolumn{1}{l}{\fourshotsecond{.863}($\uparrow$7.6\%)}                                           \\ \hline
\multicolumn{1}{c|}{\multirow{2}{*}{+ QCD \& HI}}          & \multicolumn{1}{c|}{2}                     & \twoshotsecond{.968}    & .814   & .571    & .938     & \multicolumn{1}{c|}{.730}      & \twoshotsecond{.705}     & \twoshotsecond{.960}    & \twoshotfirst{.665}     & \twoshotsecond{.843}      & \multicolumn{1}{c|}{\twoshotfirst{.729}}       & \multicolumn{1}{l}{\twoshotfirst{.792}($\uparrow$20.2\%)}                                           \\
\multicolumn{1}{c|}{}                                      & \multicolumn{1}{c|}{4}                     & \fourshotfirst{.967}    & \fourshotfirst{.977}   & .664    & \fourshotsecond{.981}     & \multicolumn{1}{c|}{.859}      & \fourshotsecond{.852}     & \fourshotfirst{.988}    & .715     & \fourshotfirst{.932}      & \multicolumn{1}{c|}{\fourshotfirst{.863}}       & \multicolumn{1}{l}{\fourshotfirst{.880}($\uparrow$9.7\%)}                                           \\ \hline
\multicolumn{12}{l|}{\textit{Others}}                                                                                                                                                                                                                   & \multicolumn{1}{l}{}                           \\ \hline
\multicolumn{1}{c|}{\multirow{2}{*}{Gemini}}               & \multicolumn{1}{c|}{2}                     & .865    & .794   & .315    & .517     & \multicolumn{1}{c|}{.704}      & .555     & .583    & .360     & .725      & \multicolumn{1}{c|}{.340}       & .576                                           \\
\multicolumn{1}{c|}{}                                      & \multicolumn{1}{c|}{4}                     & .904    & .908   & .540    & .737     & \multicolumn{1}{c|}{.861}      & .709     & .773    & .484     & .818      & \multicolumn{1}{c|}{.553}       & .729                                           \\ \hline
\multicolumn{1}{c|}{\multirow{2}{*}{SX}}               & \multicolumn{1}{c|}{2}                     & .440    & .388   & .096    & .080     & \multicolumn{1}{c|}{.060}      & .116     & .080    & .180     & .164      & \multicolumn{1}{c|}{.132}       & .174                                           \\
\multicolumn{1}{c|}{}                                      & \multicolumn{1}{c|}{4}                     & -       & -      & -       & -        & \multicolumn{1}{c|}{-}         & -        & -       & -        & -         & \multicolumn{1}{c|}{-}          & -                                              \\ \hline
\multicolumn{1}{c|}{\multirow{2}{*}{SX-IGC}} & \multicolumn{1}{c|}{2}                     & \twoshotfirst{.984}    & .568   & \twoshotfirst{.968}    & \twoshotfirst{1.00}     & \multicolumn{1}{c|}{.640}      & .516     & \twoshotfirst{.984}    & .592     & .712      & \multicolumn{1}{c|}{.628}       & \twoshotsecond{.760}                                           \\
\multicolumn{1}{c|}{}                                      & \multicolumn{1}{c|}{4}                     & -       & -      & -       & -        & \multicolumn{1}{c|}{-}         & -        & -       & -        & -         & \multicolumn{1}{c|}{-}          & -                                              \\ \hline
\end{tabular}
}
\caption{Text version results on CoBSAT.}
\label{tab:cobsat:text}
\end{table*}

\subsection{Comprehensive results on the Text-to-Image Fast Mini-ImageNet and Dreambench++}
Table~\ref{tab:t2i_imagenet_full} and Table~\ref{tab:full_dreambench++} summarize more comprehensive results on Text-to-Image Fast and Dreambench++ benchmark. 

In Text-to-Image Fast Mini-ImageNet, we evaluate the Base and TBDN with three different LVLMs and ablate the HI and QCD.
Besides, the results of MLLMs are also compared, demonstrating that our method achieves outstanding performance with different given samples (shot).  

In Dreambench++, Animal, Human, Object and Style denote the types of the personalized reference concept for concepts preservation.
At the same time, Realistic, Style and Imaginative denote prompt categories with increasing abstraction for prompt following. 
Each value is the score averaged over prompts in the corresponding subset.

\begin{table}[htbp]
\centering
\begin{tabular}{ccc}
\hline
\textbf{Method}         & \textbf{1-shot}       & \textbf{2-shot}       \\ \hline
\multicolumn{3}{l}{\textit{MLLMs}}           \\ \hline
GILL           & 16.00 ± 2.27 & 15.17 ± 2.72 \\
SL-8B  & 15.00 ± 3.27 & 12.67 ± 1.18 \\
SL-14B & 17.25 ± 2.75 & 16.75 ± 1.75 \\
Emu1       & 31.50 ± 1.87 & 22.83 ± 2.72 \\
Emu2       & 24.33 ± 3.30 & 30.67 ± 1.31 \\
Anole-7B       & 11.00 ± 2.86 & 7.00 ± 0.71  \\ \hline
\multicolumn{3}{l}{\textit{LVLM-FLUX}}       \\ \hline
Base~(Q2)       & 34.67 ± 4.48 & 44.83 ± 5.25 \\
+ HI           & 34.33 ± 7.29 & 48.17 ± 3.21 \\
+ QCD          & 35.83 ± 5.77 & 45.00 ± 3.12 \\
+ HI \& QCD    & 33.17 ± 2.29 & 47.50 ± 9.53 \\ \hline
Base~(Q2.5)     & 30.50 ± 1.80 & 34.67 ± 2.47 \\
+ HI           & 31.17 ± 1.04 & 34.33 ± 5.75 \\
+ QCD          & 33.00 ± 3.06 & 35.83 ± 1.15 \\
+ HI \& QCD    & 32.33 ± 2.78 & 35.67 ± 2.57 \\ \hline
Base~(I3)      & 34.50 ± 7.29 & 38.17 ± 5.48 \\
+ HI           & 36.50 ± 1.53 & 38.00 ± 2.18 \\
+ QCD          & 35.83 ± 0.76 & 34.00 ± 2.50 \\
+ HI \& QCD    & 39.00 ± 2.25 & 39.67 ± 2.47 \\ \hline
\end{tabular}
\caption{Results of different models on Text-to-Image Fast Mini-ImageNet (Accuracy \%)}
\label{tab:t2i_imagenet_full}
\end{table}

\begin{table*}[htbp]
\resizebox{\textwidth}{!}{
\begin{tabular}{c|ccccc|cccc|c}
\hline
\multirow{2}{*}{\textbf{Method}} & \multicolumn{5}{c|}{\textbf{Concept Preservation}} & \multicolumn{4}{c|}{\textbf{Prompt Following}}          & \multirow{2}{*}{\textbf{CP·PF$\uparrow$}} \\ \cline{2-10}
                                 & Animal & Human & Object & Style & Overall & Photorealistic & Style & Imaginative & Overall &                                 \\ \hline
SL                               & .436  & .315 & .288  & .381 & .358            & .306          & .202 & .154       & .218            & .078                           \\
SL-IGC                           & .399  & .290 & .271  & .318 & .325            & .348          & .355 & .210       & .310            & .101                           \\
SX                               & \textbf{.647}  & \textbf{.420} & \textbf{.526}  & \textbf{.571} & \textbf{.559}           & .346          & .342 & .303       & .337            & .188                           \\
SX-IGC                           & .549  & \underline{.410} & .403  & .432 & \underline{.458}            & \textbf{.922}          & \textbf{.851} & \textbf{.846}       & \textbf{.881}            & \textbf{.403}                           \\ \hline
\multicolumn{1}{l|}{TBDN(Q2)}    & \underline{.550}  & .221 & \underline{.447}  & \underline{.407} & .442            & \underline{.801}          & \underline{.799} & \underline{.698}       & \underline{.778}            & \underline{.344}                           \\ \hline
\end{tabular}
}
\caption{Comprehensive results on the Dreambench++ benchmark.}
\label{tab:full_dreambench++}
\end{table*}

\section{Qualitative results}
\label{sec:more_qualitative_results}
\subsection{Qualitative ablation on CoBSAT samples}
\begin{figure*}[htbp]
\centering
\includegraphics[width=\textwidth, height=0.8\textheight]{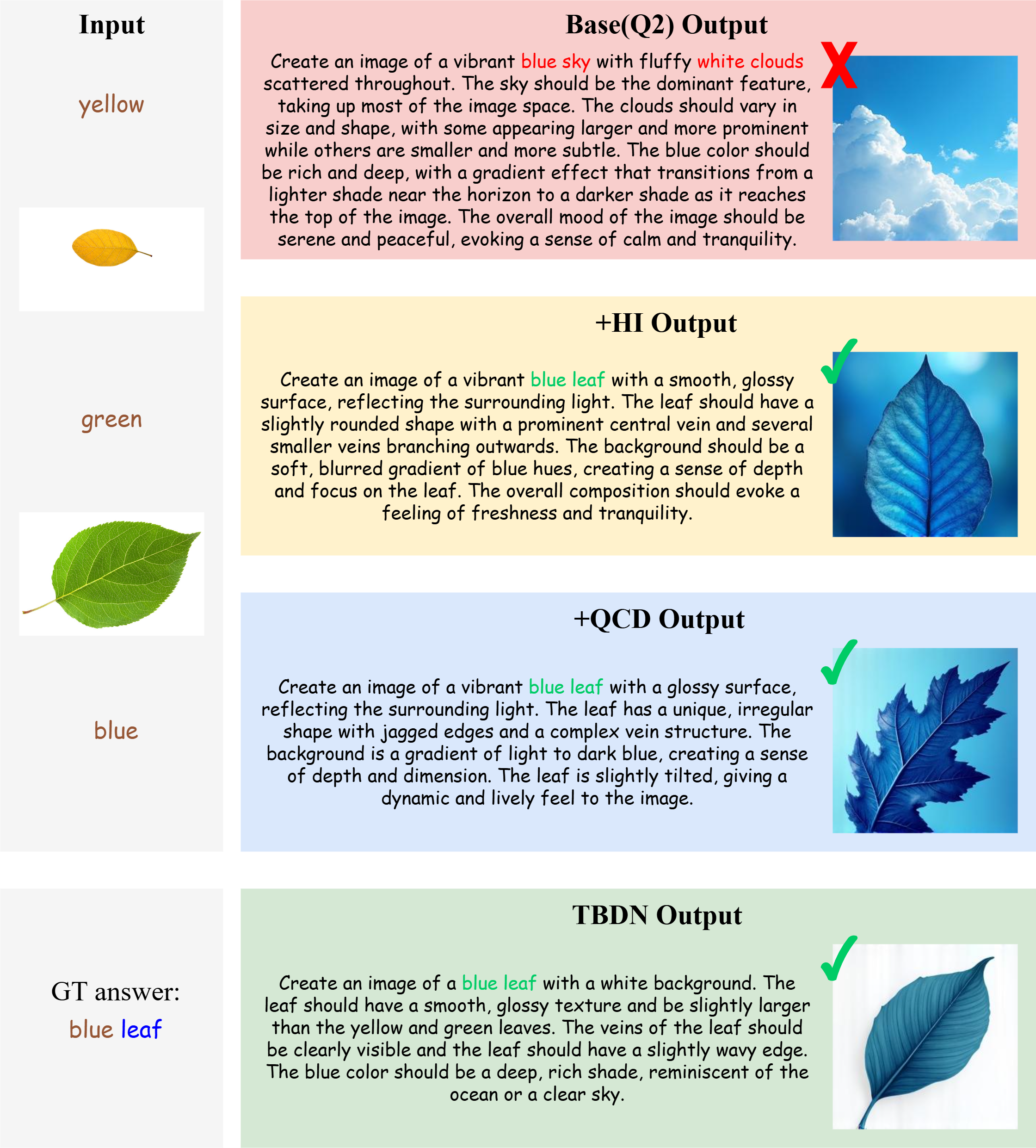}
  \caption{Qualitative results for Base (Q2), HI, QCD, and TBDN in CoBSAT (Sample 1).}
  \label{fig:cobsat_abla1}
\end{figure*}

\begin{figure*}[htbp]
\centering
\includegraphics[width=\textwidth, height=0.8\textheight]{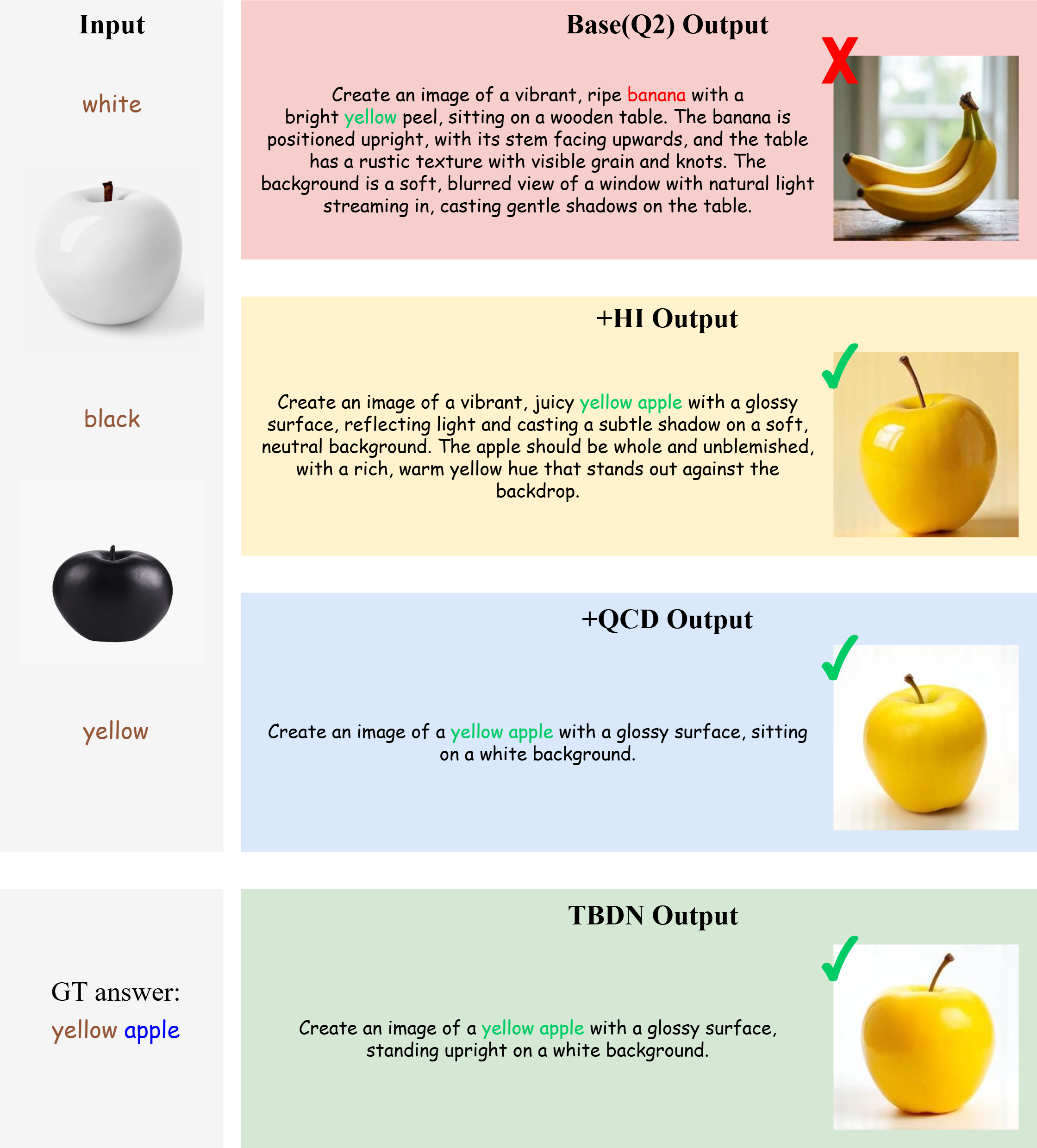}
  \caption{Qualitative results for Base (Q2), HI, QCD, and TBDN in CoBSAT (Sample 2).}
  \label{fig:cobsat_abla2}
\end{figure*}

\begin{figure*}[htbp]
\centering
\includegraphics[width=\textwidth, height=0.8\textheight]{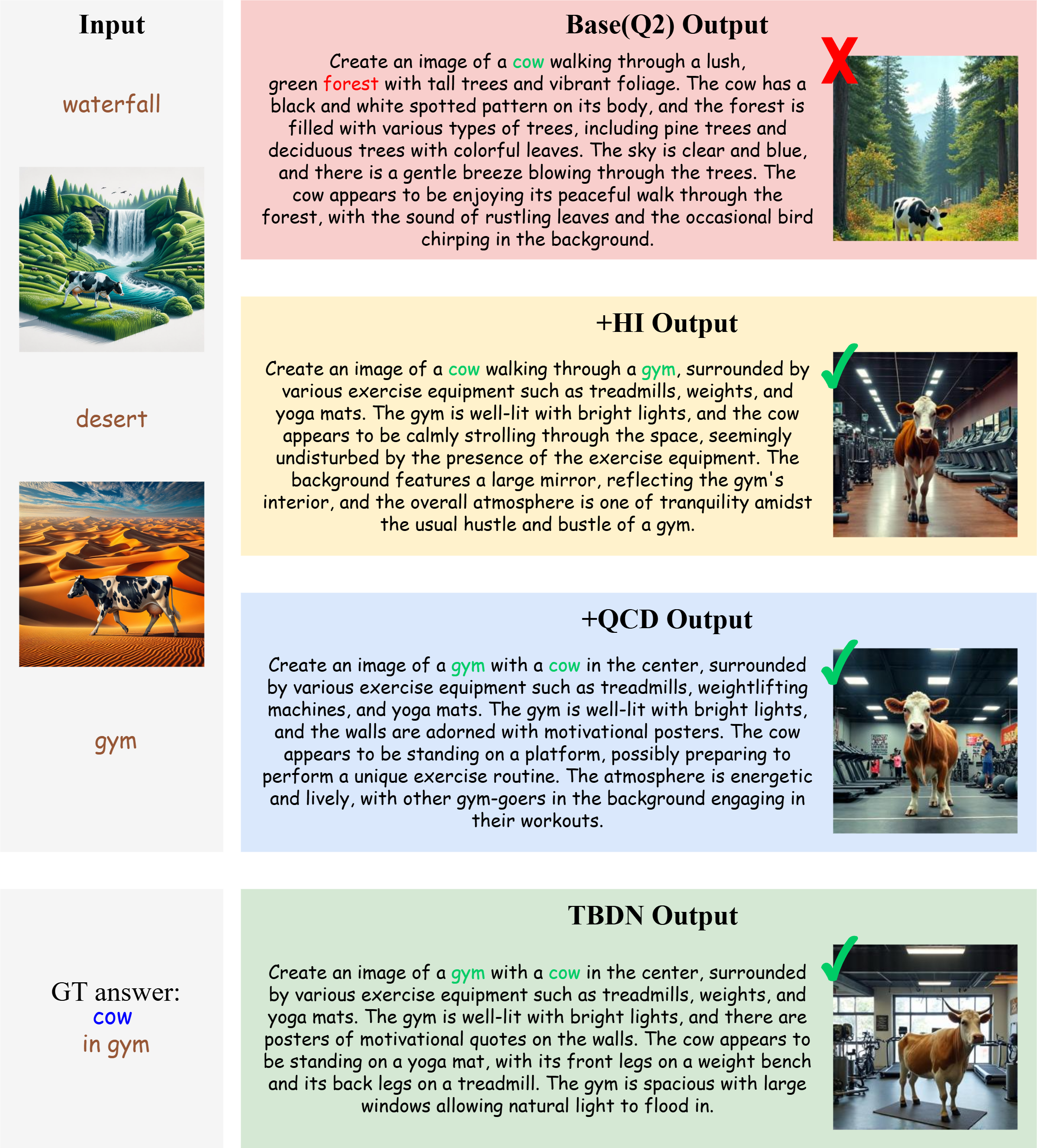}
  \caption{Qualitative results for Base (Q2), HI, QCD, and TBDN in CoBSAT (Sample 3).}
  \label{fig:cobsat_abla3}
\end{figure*}

\begin{figure*}[htbp]
\centering
\includegraphics[width=\textwidth, height=0.8\textheight]{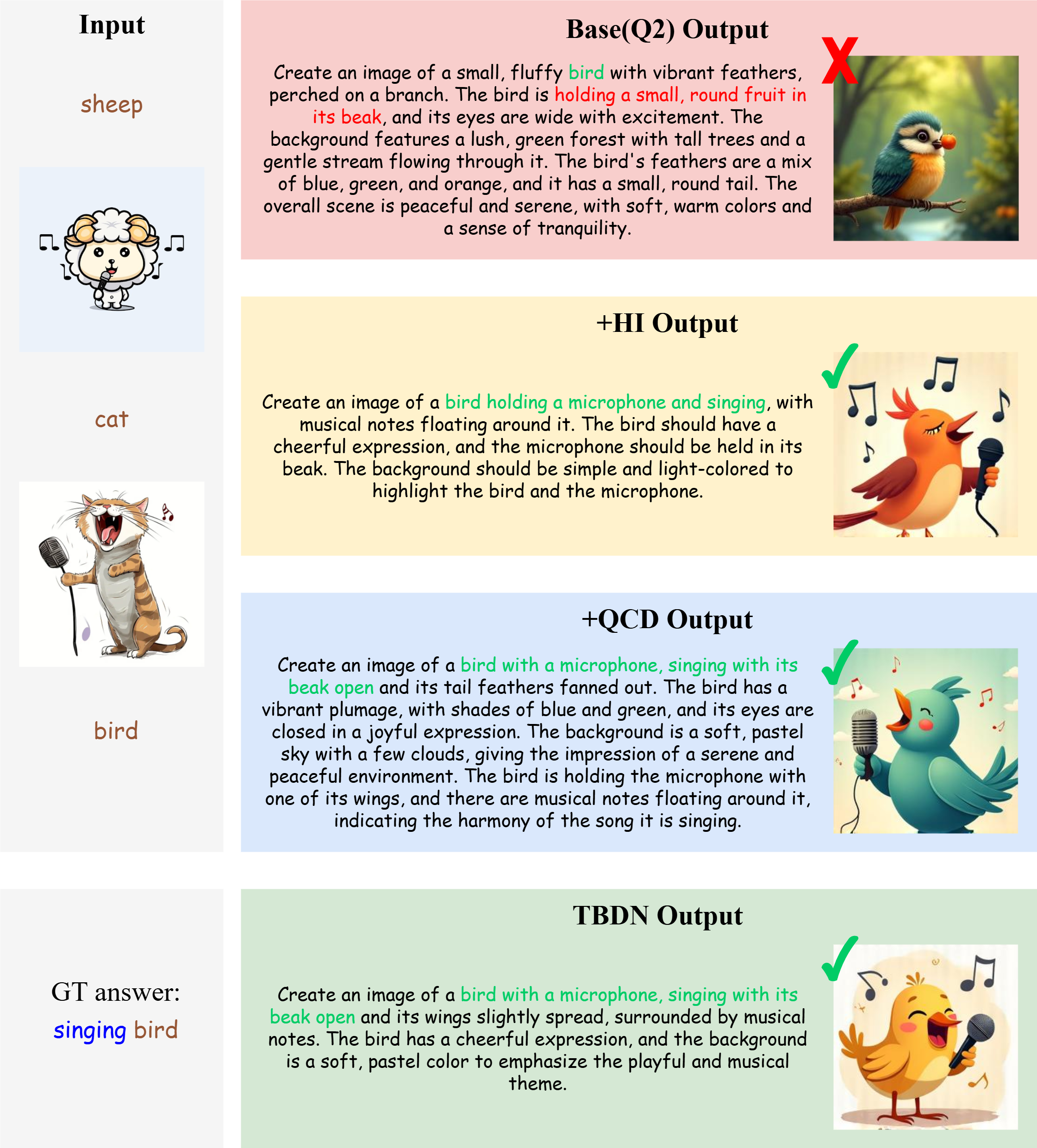}
  \caption{Qualitative results for Base (Q2), HI, QCD, and TBDN in CoBSAT (Sample 4).}
  \label{fig:cobsat_abla4}
\end{figure*}

Figure \ref{fig:cobsat_abla1}, \ref{fig:cobsat_abla2}, \ref{fig:cobsat_abla3}, \ref{fig:cobsat_abla4} show the qualitative ablation results on four samples in CoBSAT dataset.
In these cases, the Base method (see Sec.~\ref{sec:main_result} ) exhibits prior-dominated hallucinations, such as generating "blue sky", "yellow banana", "a cow in the forest", and "a bird in the forest and eating fruits".
Such text answers violate the multimodal inputs causing wrong image generation.
By explicitly injecting inductive bias, HI encourages the model to attend to the query, thereby producing correct textual outputs. 
Meanwhile, QCD mitigates the influence of priors by adjusting the output distribution. 
TBDN combines the strengths of both components and yields correct results. 
These results justify our observation and further justify the soundness and effectiveness of our method design.

\subsection{2-shot evaluation on CoBSAT}
\label{subsection:cobsat_results}
\begin{figure*}[htbp]
\centering
\includegraphics[width=\textwidth, height=0.9\textheight]{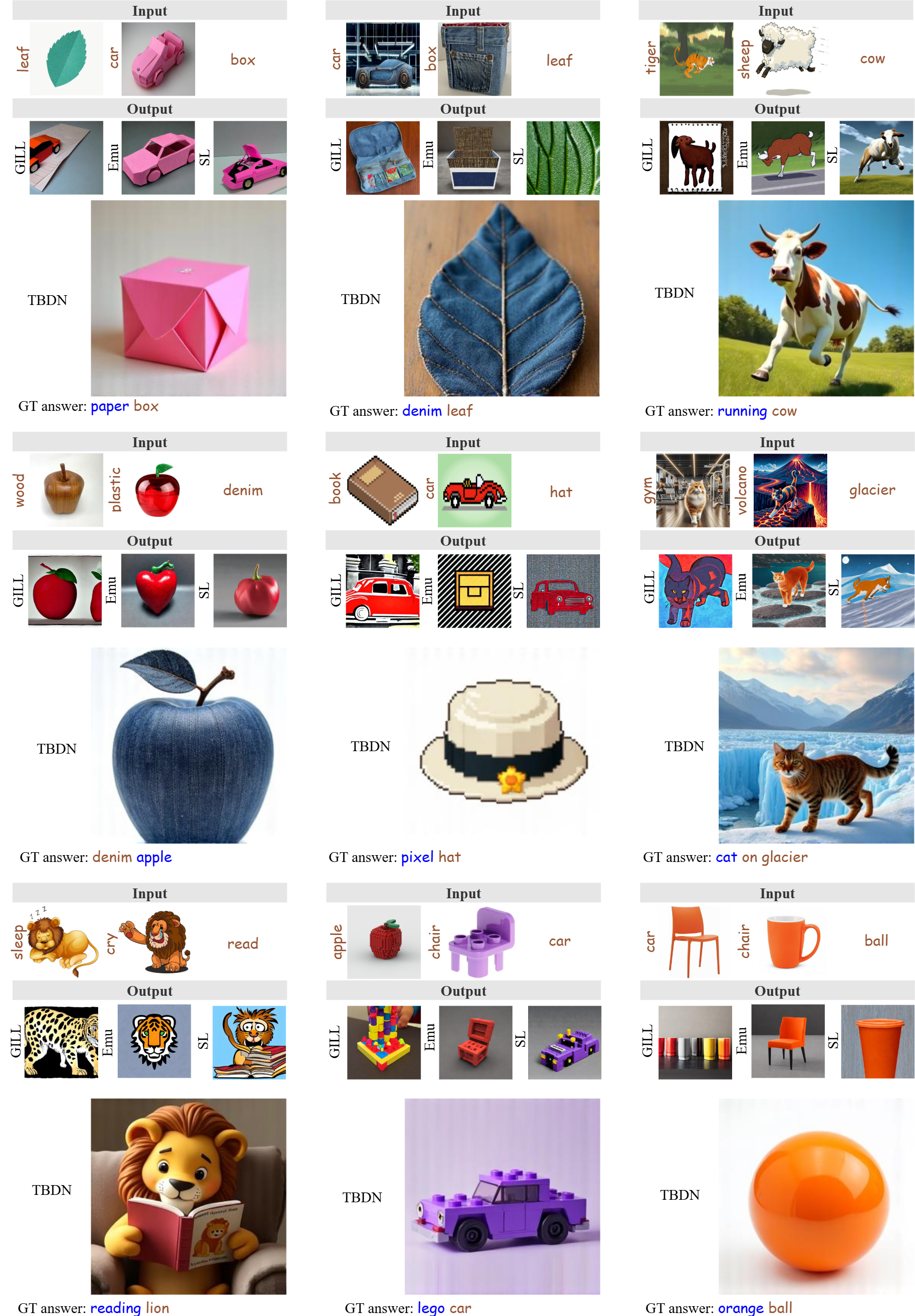}
  \caption{2-shot image generation results of TBDN and other methods on the CoBSAT benchmark.}
  \label{fig:cobsat_benchmark}
\end{figure*}

\begin{figure*}[htbp]
\centering
\includegraphics[width=\textwidth, height=0.9\textheight]{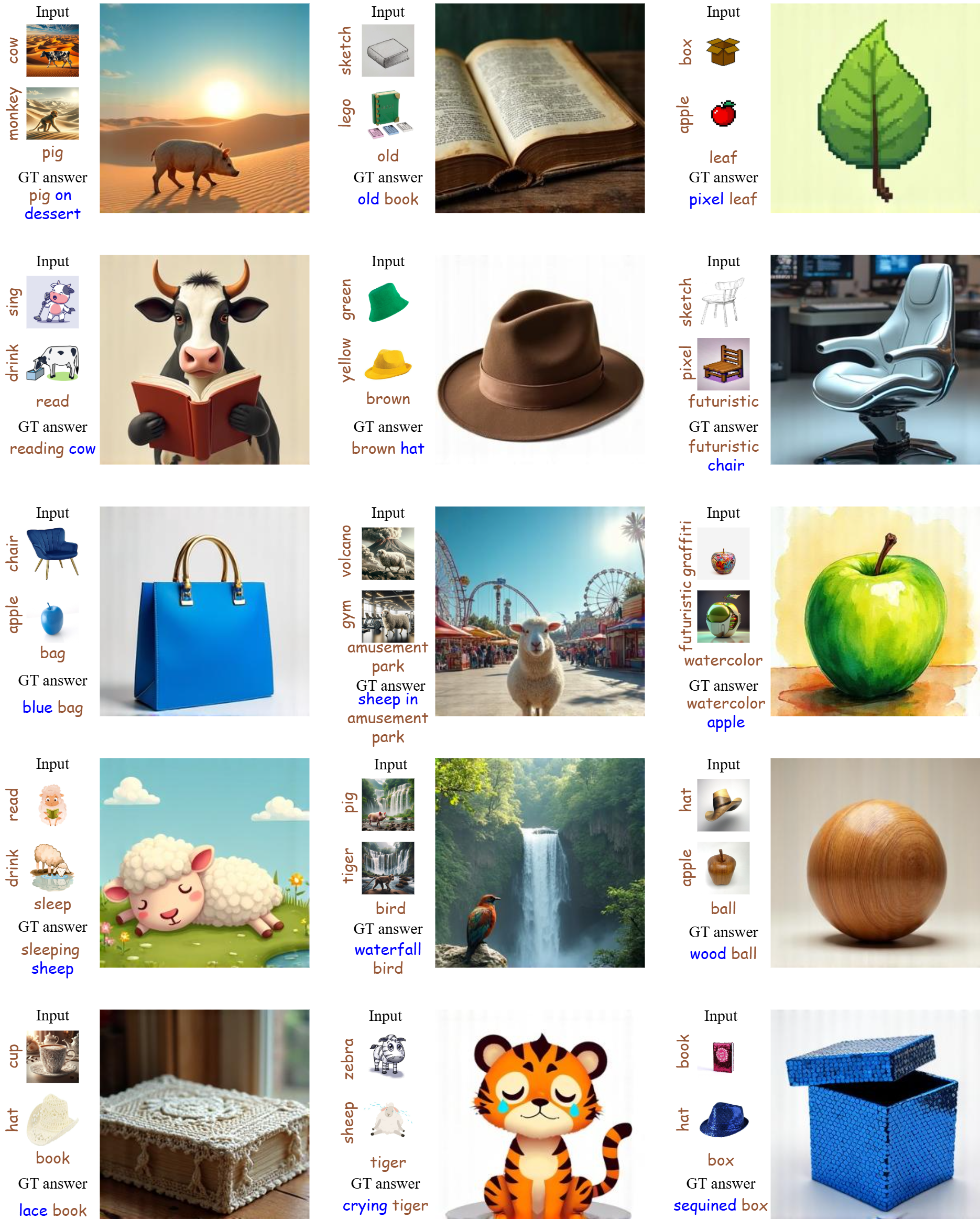}
  \caption{More 2-shot image generation results of TBDN on the CoBSAT benchmark.}
  \label{fig:cobsat}
\end{figure*}

Figure~\ref{fig:cobsat_benchmark} and Figure~\ref{fig:cobsat} show qualitative results for 2-shot evaluation on the CoBSAT benchmark. 
For each ground-truth answer, the information that must be inferred from the multimodal inputs is highlighted in \textcolor[HTML]{0000FF}{blue}, while the information explicitly provided is highlighted in \textcolor[HTML]{965635}{brown}. 
As shown in Figure~\ref{fig:cobsat}, TBDN accurately infers both the target style and subject from the multimodal context across diverse subtasks, thereby generating correct images. 
We further compare TBDN with GILL~\cite{gill}, SL~\cite{seed_llama}, and Emu1~\cite{emu} in Figure~\ref{fig:cobsat}. 
These MLLMs, to varying degrees, exhibit compliance failures (e.g., the paper box case) and prior-dominated hallucinations (e.g., the denim leaf case), whereas TBDN mitigates these issues and produces images consistent with the ground-truth answers. 
Moreover, leveraging FLUX’s~\cite{flux} strong T2I capability, FLUX can faithfully render the text produced by TBDN and generate higher-quality images than the compared MLLMs.

\subsection{Qualitative results on Text-to-Image Fast Mini-ImageNet}
\label{subsection:t2ifmi}
\begin{figure*}[htbp]
  \centering
\includegraphics[width=\textwidth, height=0.9\textheight]{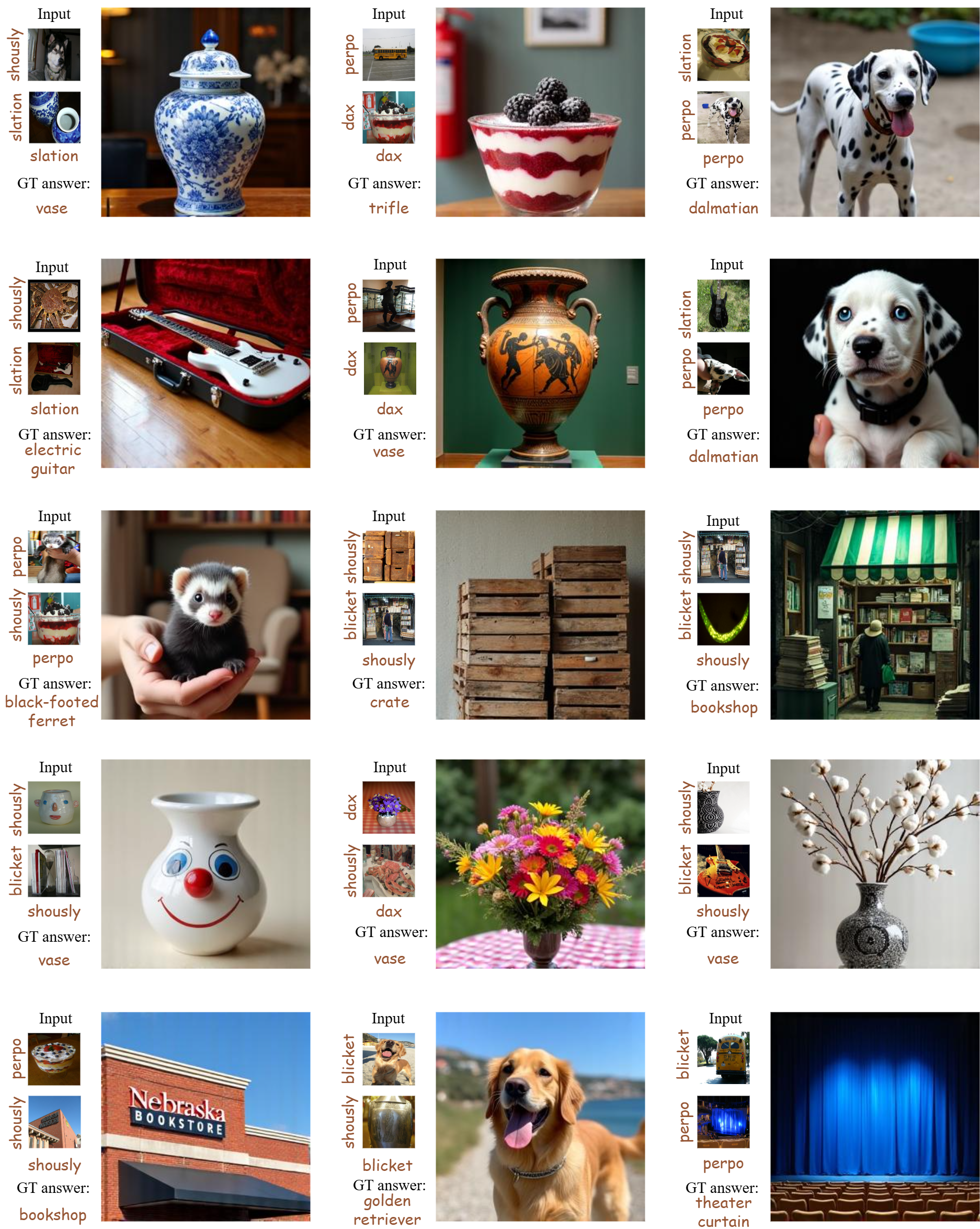}
  \caption{Image generation results on Fast Mini-ImageNet dataset.}
  \label{fig:miniimagenet1}
\end{figure*}

Figure \ref{fig:miniimagenet1} present the qualitative results of TBDN on 1-shot evaluation in T2IFMI benchmark.
TBDN can well capture the mapping between different image categories and the synthetic category name.
TBDN can then infer and generate images that closely match the queried category, yielding outputs with highly consistent subjects and styles.

\clearpage

\clearpage

\clearpage

\clearpage

\clearpage

\clearpage

\clearpage

\end{document}